\documentclass{article}

\usepackage{arxiv}

\usepackage[utf8]{inputenc} 
\usepackage[T1]{fontenc}    
\usepackage{hyperref}       
\usepackage{url}            
\usepackage{booktabs}       
\usepackage{nicefrac}       
\usepackage{microtype}      
\usepackage{lipsum}		
\usepackage{graphicx}
\usepackage{doi}

\usepackage{amsmath}
\usepackage{amssymb}

\def\real{{\mathbb R}}

\def\cL{{\mathcal L}}

\def\cR{{\mathcal R}}

\def\balpha{{\boldsymbol{\alpha}}}

\def\btheta{{\boldsymbol{\theta}}}

\def\bc{{\mathbf c}}
\def\bd{{\mathbf d}}

\def\bff{{\mathbf f}}

\def\bh{{\mathbf h}}

\def\bu{{\mathbf u}}

\def\by{{\mathbf y}}

\title{Physics-informed attention-based neural network for solving non-linear partial differential equations}

\author{Ruben Rodriguez-Torrado\thanks{Competing interest statement:OriGen AI, Inc have filed a patent application based on the research presented in this paper.} \\
	OriGen.AI and Universidad Politecnica de Madrid\\
	\texttt{rubentorrado@origen.ai} \\
	\And
	Pablo Ruiz \\
	OriGen.AI\\
	\And
	Luis Cueto-Felgueroso \\
	Universidad Politecnica de Madrid\\
	\And
	Michael Cerny Green \\
	OriGen.AI \\
	\And
	Tyler Friesen \\
	OriGen.AI \\
    \And
    Sebastien Matringe\\
    Hess Corporation\\
    \And
    Julian Togelius\\
    OriGen.AI
}

\date{}

\renewcommand{\shorttitle}{}

\begin{document}
\maketitle

\begin{abstract}
Physics-Informed Neural Networks (PINNs) have enabled significant improvements in modelling physical processes described by partial differential equations (PDEs). PINNs are based on simple architectures, and learn the behavior of complex physical systems by optimizing the network parameters to minimize the residual of the underlying PDE. Current network architectures share some of the limitations of classical numerical discretization schemes when applied to non-linear differential equations in continuum mechanics. A paradigmatic example is the solution of hyperbolic conservation laws that develop highly localized nonlinear shock waves. Learning solutions of PDEs with dominant hyperbolic character is a challenge for current PINN approaches, which rely, like most grid-based numerical schemes, on adding artificial dissipation. Here, we address the fundamental question of which network architectures are best suited to learn the complex behavior of non-linear PDEs. We focus on network architecture rather than on residual regularization. 
Our new methodology, called Physics-Informed Attention-based Neural Networks, (PIANNs), is a combination of recurrent neural networks and attention mechanisms. The attention mechanism adapts the behavior of the deep neural network to the non-linear features of the solution, and break the current limitations of PINNs. We find that PIANNs effectively capture the shock front in a hyperbolic model problem, and are capable of providing high-quality solutions inside and beyond the training set. 
\end{abstract}


\section{Introduction}\label{sec:intro}
Deep neural networks (DNNs) have achieved enormous success in recent years because they have significantly expanded the scope of possible tasks that they can perform, given sufficiently large datasets~\cite{Sejnowski_PNAS_2020}. The range of applications is extraordinary, from natural language processing~\cite{cho2014learning,sutskever2014sequence}, image analysis~\cite{hemanth2017deep} and autonomous driving~\cite{grigorescu2020survey}, to earthquake forecasting~\cite{johnson_pnas_2021}, playing videogames~\cite{torrado2020bootstrapping,torrado2018deep} and, more recently, numerical differentiation~\cite{Bar-Sinai_PNAS_2019}.

Neural networks can approximate the solution of differential equations~\cite{regazzoni_jcp_2019,samaniego_cmame_2020}, in particular high-dimensional partial differential equations (PDEs)~\cite{Han_PNAS_2018,Beck_arxiv_2021}. One of the most remarkable approaches to solve non-linear PDEs is physics-informed neural networks (PINNs)~\cite{raissi2019physics,raissi2017physics}. PINNs are trained to solve supervised learning tasks constrained by PDEs, such as the conservation laws in continuum theories of fluid and solid mechanics~\cite{Raissi_Science_2020,Brunton_Annurev_2020,kadeethum2020physics,samaniego_cmame_2020,Haghighat_CMAME_2021b}. The idea behind PINNs is to train the network using automatic differentiation (AD) by calculating and minimizing the residual, usually constrained by initial and boundary conditions, and possibly observed data~\cite{raissi2019physics}. PINNs have the potential to serve as on-demand, efficient simulators for physical processes described by differential equations (the \emph{forward} problem)~\cite{raissi2019physics,jin2021nsfnets,kadeethum2020physics}. If trained accurately, PINNs can work faster and more accurately than numerical simulators of complex real-world phenomena. PINNs may also be used to assimilate data and observations into numerical models, or be used in parameter identification (the \emph{inverse} problem)~\cite{raissi2019physics,yang2021b} and uncertainty quantification~\cite{Mao_CMAME_2020,jagtap_cmame_2020,zhang_jcp_2019,tipireddy_jcp_2020}. 

Learning solutions of nonlinear PDEs using current network architectures presents some of the same limitations of classical numerical discretization schemes. A paradigmatic example is the solution of hyperbolic PDEs. Hyperbolic conservation laws describe a plethora of physical systems in gas dynamics, acoustics, elastodynamics, optics, geophysics, and biomechanics~\cite{dafermos}. Hyperbolic PDEs are challenging to solve numerically using classical discretization schemes, because they tend to form self-sharpening, highly-localized, nonlinear shock waves that require specific approximation strategies and fine meshes~\cite{leveque}. On the other hand, the ability of current PINNs to learn PDEs with a dominant hyperbolic character relies on adding artificial dissipation~\cite{fuks2020limitations,fraces2020physics,fraces2021physics,Michoski_Neuro_2020}, or on using a priori knowledge to increase the number of training points along the shock trajectories~\cite{Mao_CMAME_2020}.

In this work, we propose a new perspective on solving nonlinear PDEs using deep learning, by focusing on network architectures rather than on residual regularization. We explore two core ideas: 1) A modified PINN architecture can provide a more general method for solving hyperbolic conservation law problems without a priori knowledge or residual regularization. 2) Relating network architecture with the physics encapsulated in a given PDE is possible and has a beneficial impact. Our hypothesis is that sophisticated, physics-specific network architectures (e.g. networks whose internal hierarchy is related to the physical processes being learned) may be more effectively trained and understood than standard feed-forward multilayer perceptrons.


The proposed new architecture, inspired by recent advances in deep learning for language processing and translation~\cite{bahdanau2014neural,vaswani2017attention}, is a combination of general recurrent units (GRUs) and \emph{attention mechanisms}; we call this a physics-informed attention-based neural network (PIANN). The combination of both elements in the architecture allows for determination of the most relevant information (recurrent neural network with memory) to adapt the behavior of the deep neural network to approximate sharp shocks without the necessity of residual regularization or a priori knowledge (attention mechanism). We use a classical hyperbolic model problem (the Buckley-Leverett equation~\cite{BL,leveque}) as a benchmark for hyperbolic conservation laws to test our methodology, and find that PIANNs effectively capture the shock front and are capable of providing high quality solutions inside and beyond the training set.

\section{Problem formulation}\label{sec:problem-formulation}

The problem of interest is that of two immiscible fluids (oil and water) flowing through a porous medium (sand). The Buckley-Leverett (BL) equation~\cite{BL} describes the evolution in time and space of the wetting-phase (water) saturation.
Let $u_M:\real_0^+\times \real_0^+ \rightarrow [0,1]$ be the solution of the BL equation 
\begin{align}
&\frac{\partial u_M}{\partial t}(x,t) + \frac{\partial f_M}{\partial x} (x,t) = 0, \label{eq:pde} \\
u_M(x,0) &= 0, \forall x> 0, \text{\ \ \ Initial condition} \label{eq:initial_conditions}\\
u_M(0,t) &= 1, \forall t \geq 0, \text{ \ \ \ Boundary condition} \label{eq:boundary_condition}
\end{align}
where $u_M$ usually represents the wetting-phase saturation, $f_M$ is the fractional flow function and $M$ is the mobility ratio of the two fluid phases. 

This first-order hyperbolic equation is of interest as its solution can display both smooth solutions (rarefactions) and sharp fronts (shocks). Although the solution to this problem can be calculated analytically, the precise and stable resolution of these shocks poses well-known challenges for numerical methods~\cite{leveque}.

Physics-Informed Neural Networks (PINNs) have been tested on this problem by Fuks and Tchelepi~\cite{fuks2020limitations} who report good performance for concave fractional flow functions. The solution of the problem in the case of a non-concave fractional flow function is, however, much more challenging and remains an open problem. 
We take $f_M$ to be the non-concave flux function
\begin{equation}
f_M(x,t) = \frac{u_M(x,t)^2}{ u_M(x,t)^2 + \frac{1}{M}\left(1-u_M(x,t)\right)^2},
\end{equation}
for which we can obtain the analytical solution of the problem:
\begin{equation}\label{eq:analytical_solution}
u_M(x,t) =  \begin{cases}
0, & \frac{x}{t} > f'_{M}(u^*), \\ 
u(x/t), & f'_M(u^*) \geq \frac{x}{t}\geq f'_M(u=1), \\
1, & f'_M(u=1) \geq \frac{x}{t},
\end{cases}
\end{equation}
where $u^*$ represents the shock location defined by the Rankine-Hugoniot condition~\cite{leveque}.

\section{Methodology}\label{sec:methodology}

Let ${\cal G} := \left\{ (x_i,t_j) \in \real_0^+\times \real_0^+ : i=0,\hdots, N, j=0,\hdots,T\right\} $ be a discrete version of the domain of $u_M$. We define our PIANN as a vector function $\bu_\btheta:\real_0^+ \times \real^+ \rightarrow [0,1]^{N+1}$, where $\btheta$ are the weights of the network to be estimated during training.
The inputs for the proposed architecture are pairs of $(t,M)$ and the output is a vector where the $i$-th component is the solution evaluated in $x_i$. Notice the different treatment applied to spatial and temporal coordinates. Whereas $t$ is a variable of the vector function $\bu_{\btheta}$, the locations where we calculated the solution $x_0,\hdots,x_N$ are fixed in advance. The output is a saturation map and therefore its values have to be in the interval $[0,1]$.

For the sake of readability, we introduce the architecture of $\bu_{\btheta}$ in section \ref{sec:pinn_architecture}. However, we advance that in order to enforce the boundary condition, we let our PIANN learn only the components $\bu_{\btheta}(t,M)_1,\hdots,\bu_{\btheta}(t,M)_{N}, \forall t\neq 0$ and then we concatenate the component $\bu_{\btheta}(t,M)_0 = 1$.
To enforce the initial conditions, we set $\bu_{\btheta}(0,M)_i = 0, i=1, \hdots,N$. To enforce that the solution be in the interval $[0,1]$, a sigmoid activation function is applied to each component of the last layer of our PIANN.


The parameters of the PIANN are estimated according to the physics-informed learning approach, which states that $\btheta$ can be estimated from the BL equation eq.~(\ref{eq:pde}), the initial conditions eq.~(\ref{eq:initial_conditions}) and boundary conditions eq.~(\ref{eq:boundary_condition}), or in other words, no examples of the solution are needed to train a PINN. 

After utilizing the information provided by the initial and boundary conditions enforcing $\bu_{\btheta}(0,M)_i = 0, i=1,\hdots,N$ and $\bu_{\btheta}(t,M)_0 = 1$, respectively, we now define a loss function based on the information provided by eq.~(\ref{eq:pde}). To calculate the first term we propose two options. The first option is a central finite difference approximation, that is,
\begin{equation}
\cR_1(\btheta,M)_{i,j} = \frac{\bu_{\btheta}(t_{j+1},M)_i - \bu_{\btheta}(t_{j},M)_i}{t_{j+1} - t_{j}},\begin{array}{c} \scriptstyle i=1,\hdots,N-1\\ \scriptstyle j=1,\hdots,T-1\end{array}.
\end{equation}
Alternatively, we can calculate the derivative of our PIANN with respect to $t$ since we know the functional form of $\bu_{\btheta}$. It can be calculated using the automatic differentiation tools included in many machine learning libraries, such as Pytorch. Thus, we propose a second option to calculate this term as $\cR_1(\btheta,M)_{i,j} = \partial \bu_{\btheta}(t,M)_i / \partial t |_{t = t_j}$.

The second term of eq.~(\ref{eq:pde}), the derivative of the flux with respect to the spatial coordinate, is approximated using central finite difference as
\begin{equation}
\cR_2(\btheta,M)_{i,j} = \frac{\bff(t_j,M)_{i+1} - \bff(t_j,M)_{i-1}}{x_{i+1} - x_{i-1}}, \begin{array}{c} \scriptstyle i=1,\hdots,N-1\\ \scriptstyle j=1,\hdots,T-1\end{array},
\end{equation}
where the vector of fluxes at the $i$-th location $x_i$ is calculated as
\begin{equation}
\bff_{\btheta}(t,M)_i = \frac{\bu_{\btheta}(t,M)_i^2}{\bu_{\btheta}(t,M)_i^2 - \frac{\left(1-\bu_{\btheta}(t,M)_i\right)^2}{M}}, i=0, \hdots N.
\end{equation}
The spatial coordinate $x$ is included as a fixed parameter in our architecture. 

The loss function to estimate the parameters of the PINN is given as 

\begin{equation}\label{eq:loss}
\cL(\btheta) = \sum_{M} \left\| \cR_1(\btheta,M) + \cR_2(\btheta,M) \right\|_F^2,
\end{equation}
where $\|\cdot\|_F$ is the Fröbenius norm.

It should be noted that unlike all previous physics-informed learning works in the literature (recall section \ref{sec:intro}), the initial and boundary conditions are not included in the loss function; they are already enforced in the architecture. This has three direct consequences. First, we are enforcing a stronger constraint that does not allow any error on the initial and boundary conditions. Second, the PIANN does not need to learn these conditions by itself, and it can concentrate only on learning the parameters that minimize the residuals of the BL equation. Third, since we only have the term of the residuals, there are no weights to be tuned to control the effect of the initial and boundary conditions in the final solution.

Finally, the parameters of the PIANN are estimated using ADAM optimizer~\cite{kingma2014adam} to minimize eq.~(\ref{eq:loss}) with respect to $\btheta$.

\section{PIANN architecture}\label{sec:pinn_architecture}

Although it has been demonstrated that neural networks are universal function approximators, certain challenging problems (e.g. solving non-linear PDEs) may require more specific architectures to capture all their properties. For that reason, we have proposed a new architecture, inspired by~\cite{bahdanau2014neural}, to solve non-linear PDEs with discontinuities under two assumptions.

First, to automatically detect discontinuities we need an architecture that can exploit the correlations between the values of the solution for all spatial locations $x_1,\hdots,x_N$. Second, the architecture has to be flexible enough to capture different behaviors of the solution at different regions of the domain. To this end, we propose the use of encoder-decoder GRUs~\cite{cho2014learning} for predicting the solution at all locations at once, with the use of a recent machine learning tool known as attention mechanisms~\cite{vaswani2017attention}.  

Our approach presents several advantages compared to traditional simulators: i) Instead of using just neighboring cells' information to calculate $\bu$ as in numerical methods, our architecture uses the complete encoded sequence input of the grid to obtain $\bu_{i}$, allowing us to capture non-local relationships that numerical methods struggle to identify. ii) the computer time for the forward pass of neural networks models is linear with respect to the number of cells in our grid. In other words, our method is a faster alternative with respect to traditional methods of solving PDEs.





\begin{figure*}
\centering
\includegraphics[width=0.8\linewidth]{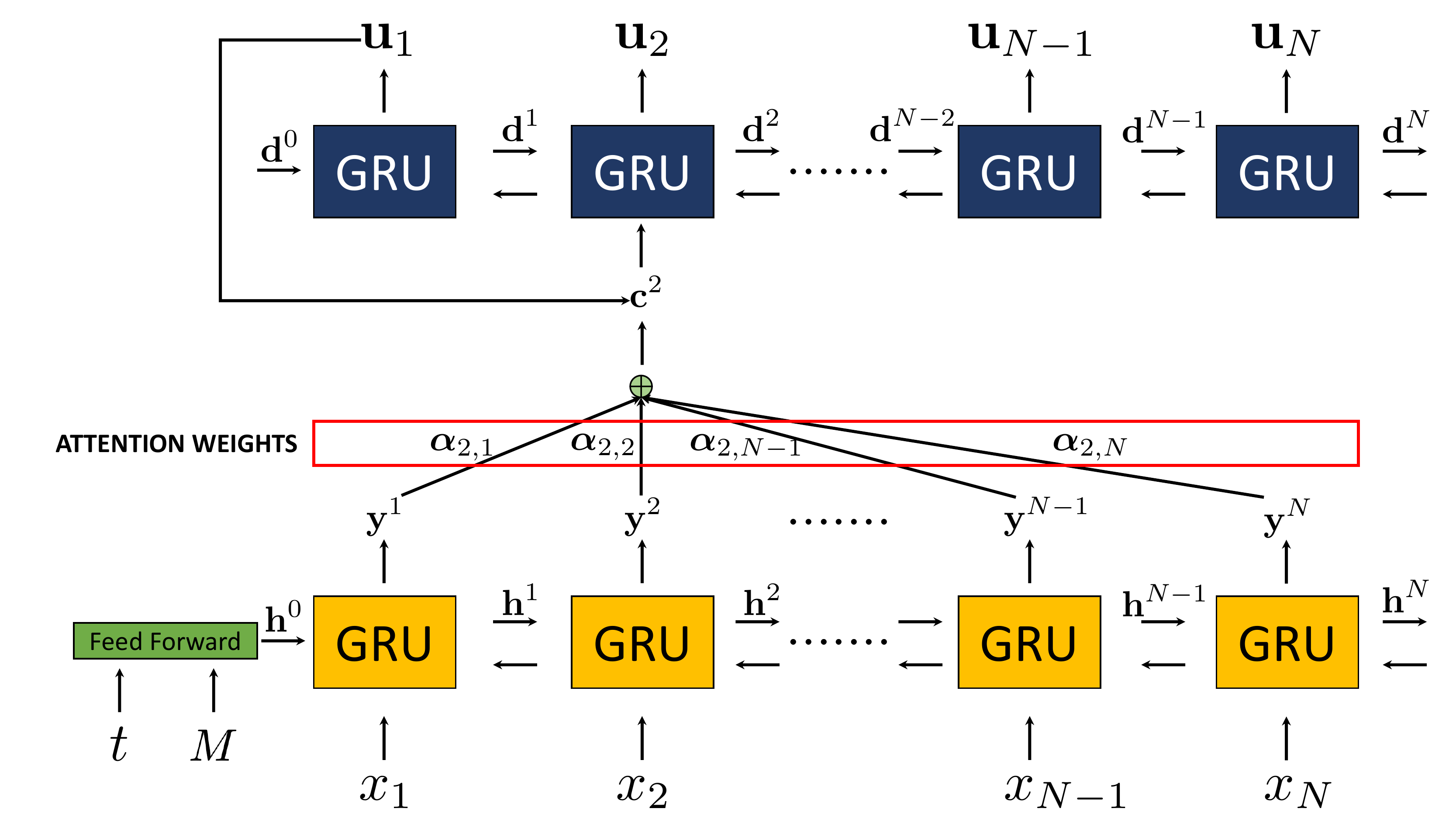}
\caption{Architecture of physical attention neural network for the prediction of the variable $\bu_{2}$}
\label{fig:architecture}
\end{figure*}

Figure \ref{fig:architecture} shows an outline of the proposed architecture. We start feeding the input pair $(t,M)$ to a single fully connected layer. 
Thus, we obtain $\bh^0$ the initial hidden state of a sequence of $N$ GRU blocks (yellow). Each of them corresponds to a spatial coordinate $x_i$ which is combined with the previous hidden state $\bh^{i-1}$ inside the block.

This generates a set of vectors $\by^1,\hdots,\by^N$ which can be understood as a representation of the input in a latent space. The definitive solution $\bu$ (we omit the subindex $\btheta$ for simplicity) is reached after a new sequence of GRU blocks (blue) whose initial hidden state $\bd^0$ is initialized as $\bh^N$ to preserve the memory of the system.

In addition to the hidden state $\bd^i$, the $i$-th block $g_i$ is fed with a concatenation of the solution at the previous location and a context vector, that is

\begin{equation}
    \bu_{i} = g_i([\bu_{i-1},\bc^i],\bd^{i-1}).
\end{equation}

How the context vector is obtained is one of the key aspects of our architecture, since it will provide the PINN with enough flexibility to fit to the different behaviors of the solution depending on the region of the domain. Inspired by~\cite{bahdanau2014neural}, we introduce an attention mechanism between both GRU block sequences. Our attention mechanism is a single fully connected layer, $a$, that learns the relationship between each component of $\by^j$ and the hidden states of the (blue) GRU sequence,

\begin{equation}
{\cal E}_{i,j} = a(\bd^{i-1},\by^j).
\end{equation}

Then, the rows of matrix ${\cal E}$ are normalized using a softmax function as
\begin{equation}
\balpha_{i,j} = \frac{\exp\left({\cal E}_{i,j}\right)}{\sum_{j=1}^N\exp\left({\cal E}_{i,j}\right)},
\end{equation}
and the context vectors are calculated as
\begin{equation}
    \bc^i = \sum_{j=1}^N \balpha_{i,j} \by^j, i =1,\dots,N.
\end{equation}

The coefficients $\balpha_{i,j}$ can be understood as the degree of influence of the component $\by^j$ in the output $\bu_{i}$. This is one of the main innovations of our work to solve hyperbolic equations with discontinuities. The attention mechanism automatically determines the most relevant encoded information of the full sequence of the input data to predict the $\bu_i$. In other words, attention mechanism is a new method that allows one to determine the location of the shock automatically and provide more accurate behavior of the PIANN model around this location. This new methodology breaks the limitations explored by other authors~\cite{mao2020physics}~\cite{fuks2020limitations}~\cite{fraces2020physics} since is is able to capture the discontinuity without specific prior information or the regularization term of the residual.   


This is the first paper to use attention mechanisms to solve non-linear PDEs for hyperbolic problems with discontinuity.

\section{Results}\label{sec:experiments}






In this section we perform a set of experiments that support the proposed methodology. The goal of our experiments is to demonstrate that our PIANN is indeed able to approximate the analytical solution given in eq.~(\ref{eq:analytical_solution}).  

The training set is given by a grid ${\cal G} =\{(x_i,t_j) \in \real_0^+\times \real_0^+ : x_i \in \{0,0.01,\hdots,0.99,1\},t_j\in\{0,0.01,\hdots, 0.49, 0.5\}\}$, and a set of values of $M \in \{2,4,6,\hdots,100\} $, which produces $N = 101$, $T = 51$, and a total of 257,550 points. We want to emphasize that no examples of the solution are known at these points, and therefore no expensive and slow simulators are required to build the training set.
To estimate the parameters of the PIANN we minimize eq.~(\ref{eq:loss}) by running ADAM optimizer for $200$ epochs with a learning rate of $0.001$.

\begin{figure}
\centering
\includegraphics[width=0.5\linewidth]{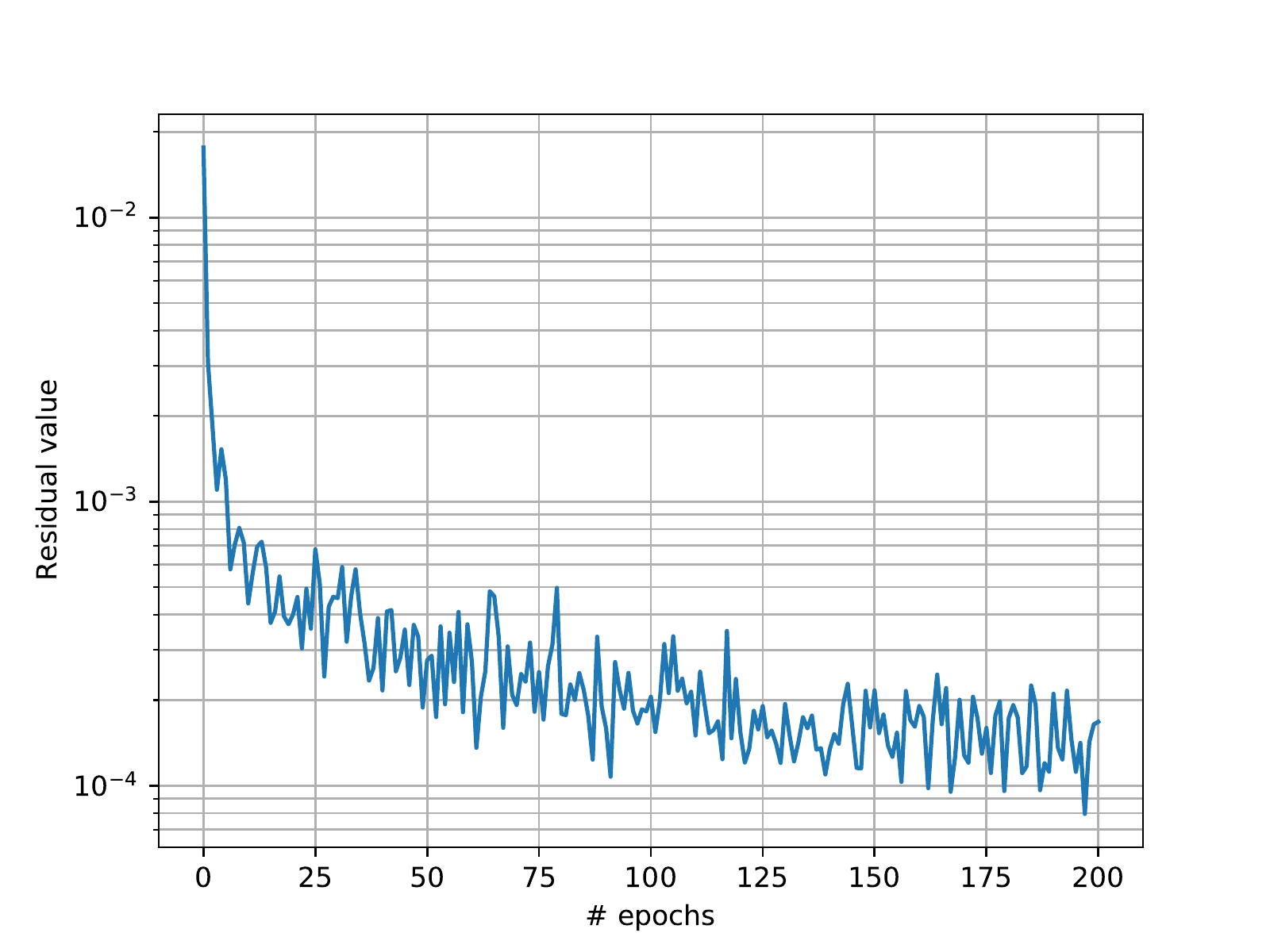}
\caption{Residual values for each epoch for $M=4.5$ values in a semilog scale.}
\label{fig:residuals}
\end{figure}

Figure \ref{fig:residuals} shows the residual value for the testing dataset for the different epoch for $M=4.5$. We can observe a fast convergence of the method  and a cumulative value of the residual smaller than $10^{-4}$ after a few epochs. This demonstrates that we are minimizing the residuals in eq.~(\ref{eq:pde}) and subsequently solving the the equation that governs BL. 

\begin{figure*}
     \centering
     \includegraphics[width=0.28\linewidth]{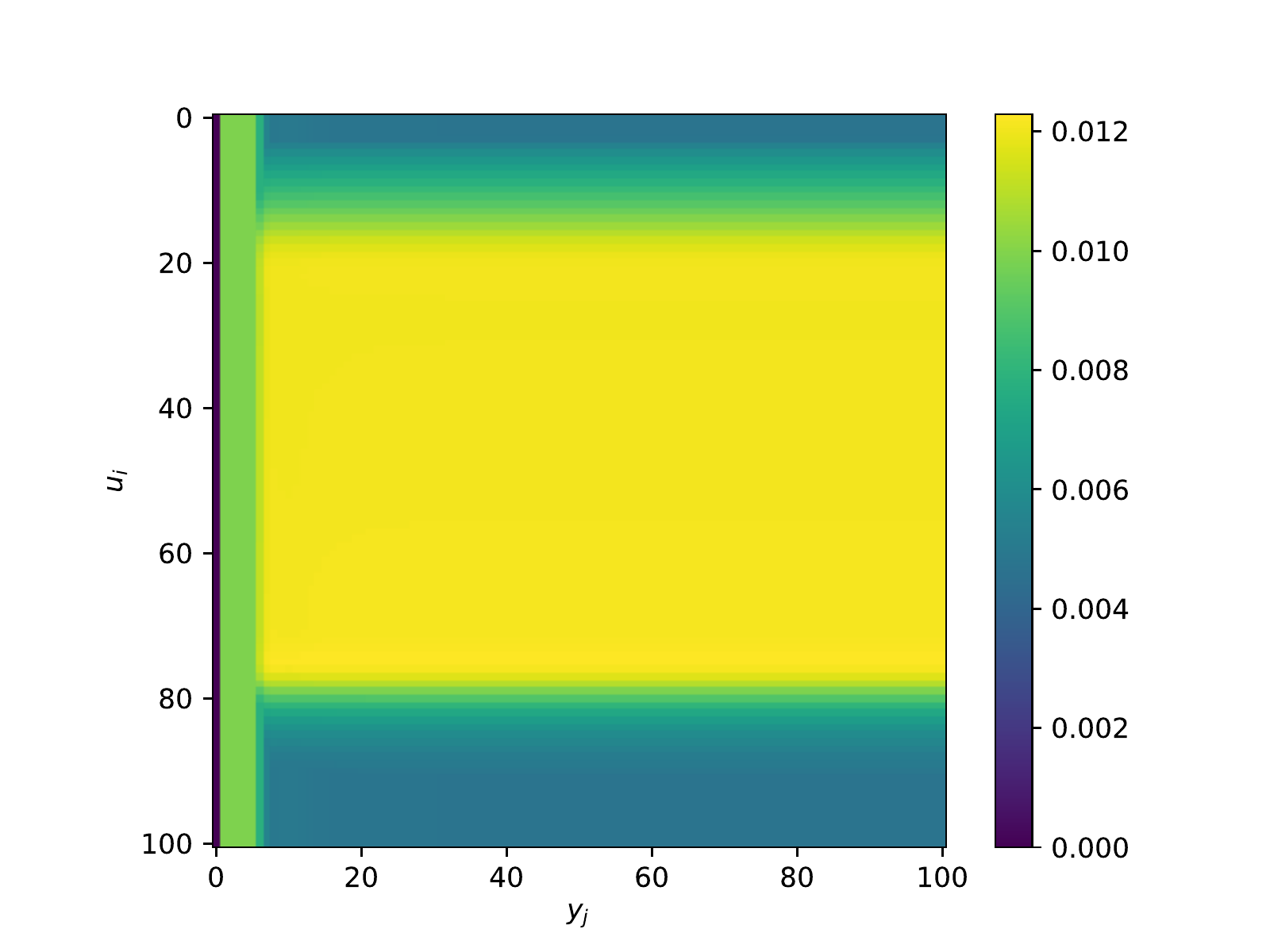}
     \includegraphics[width=0.28\linewidth]{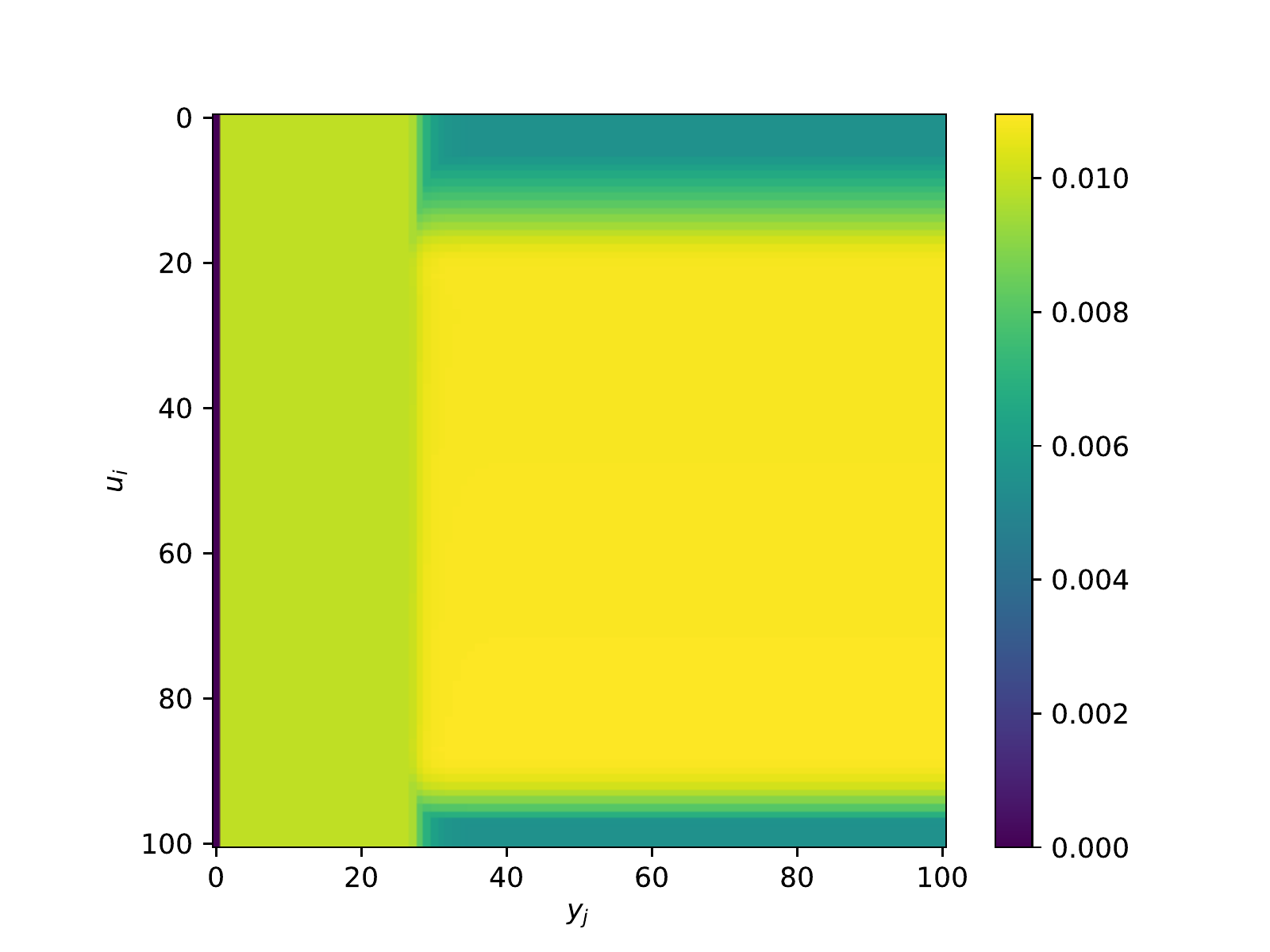}
     \includegraphics[width=0.28\linewidth]{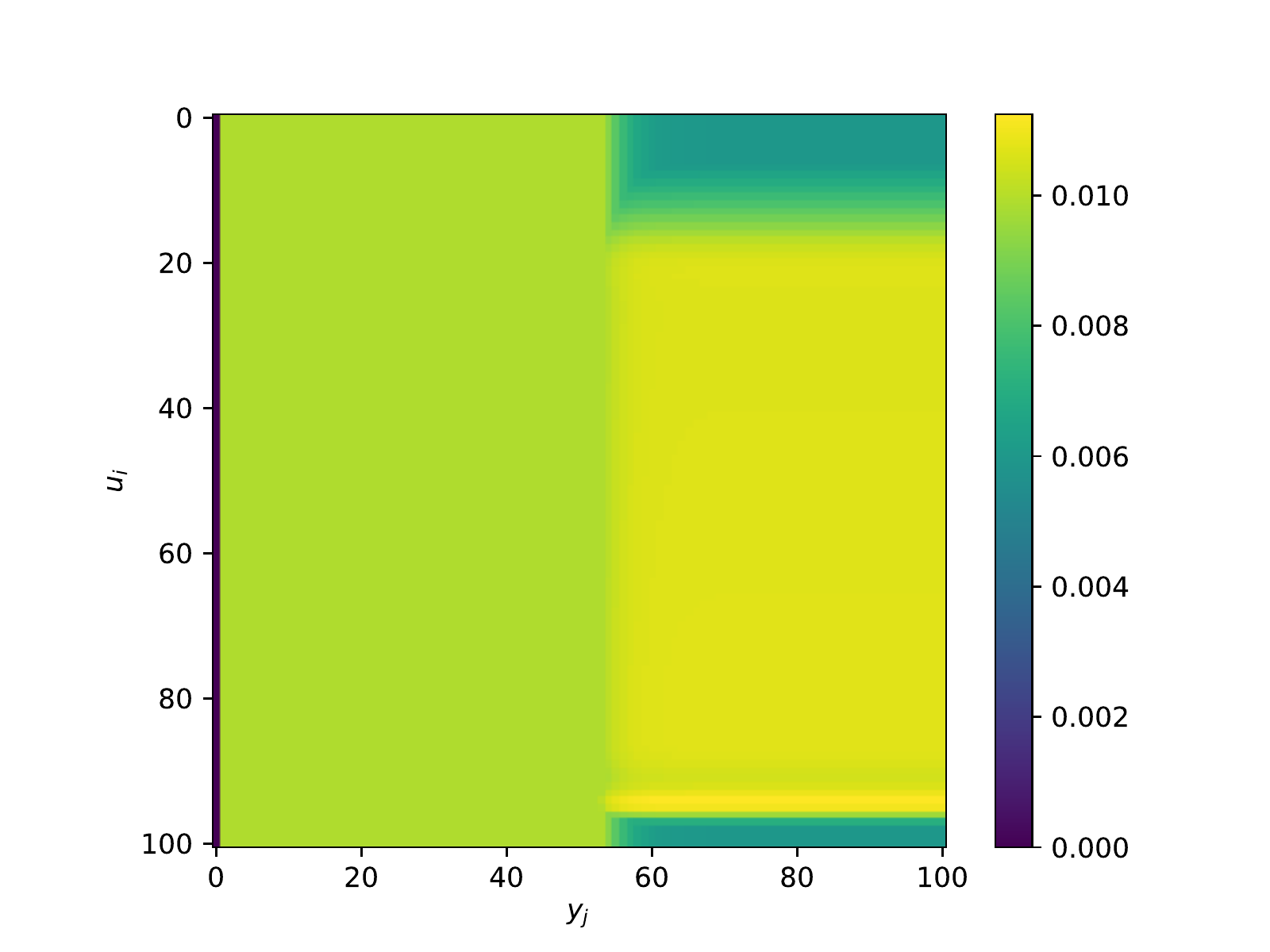}
     \includegraphics[width=0.28\linewidth]{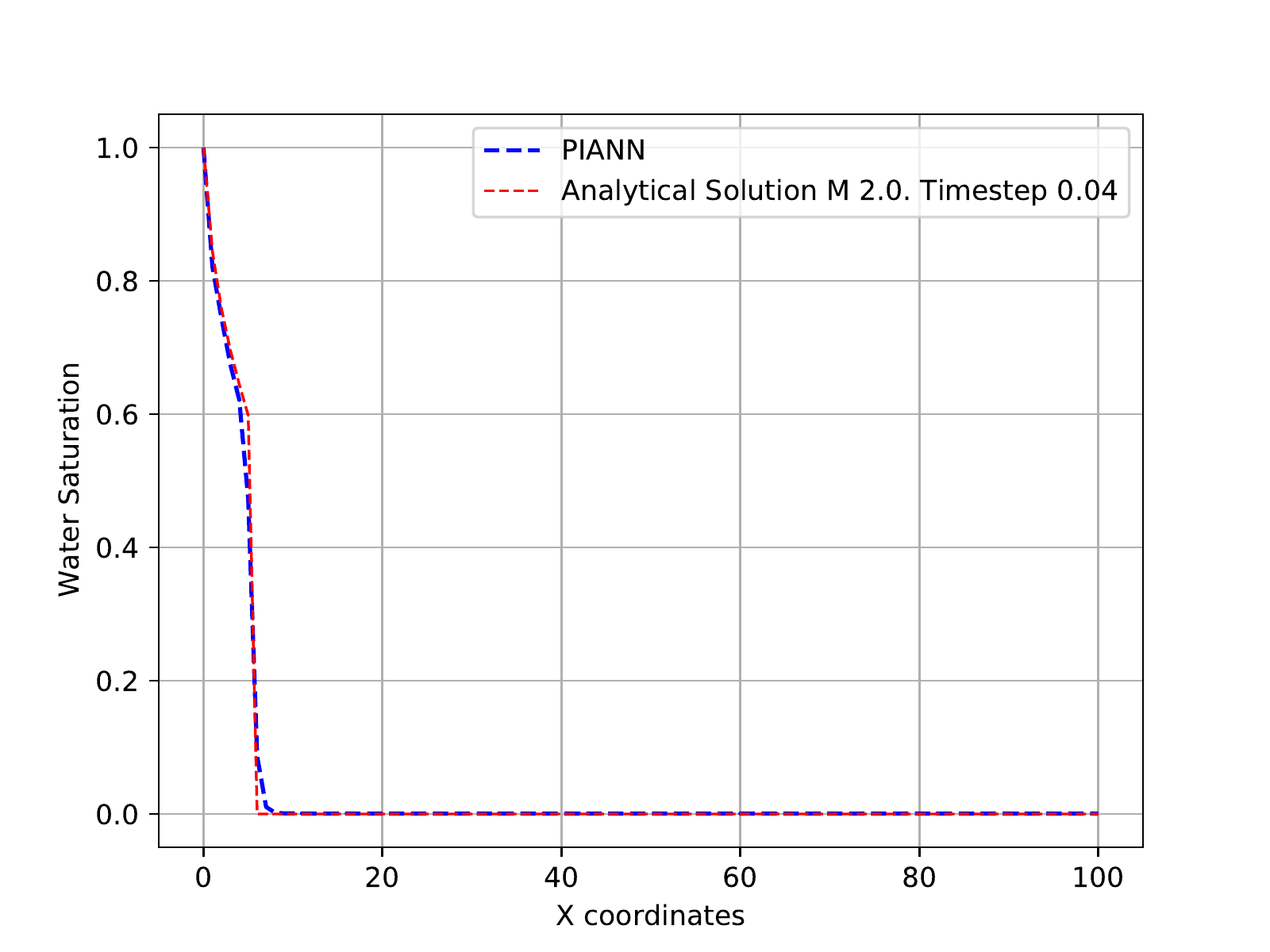}
     \includegraphics[width=0.28\linewidth]{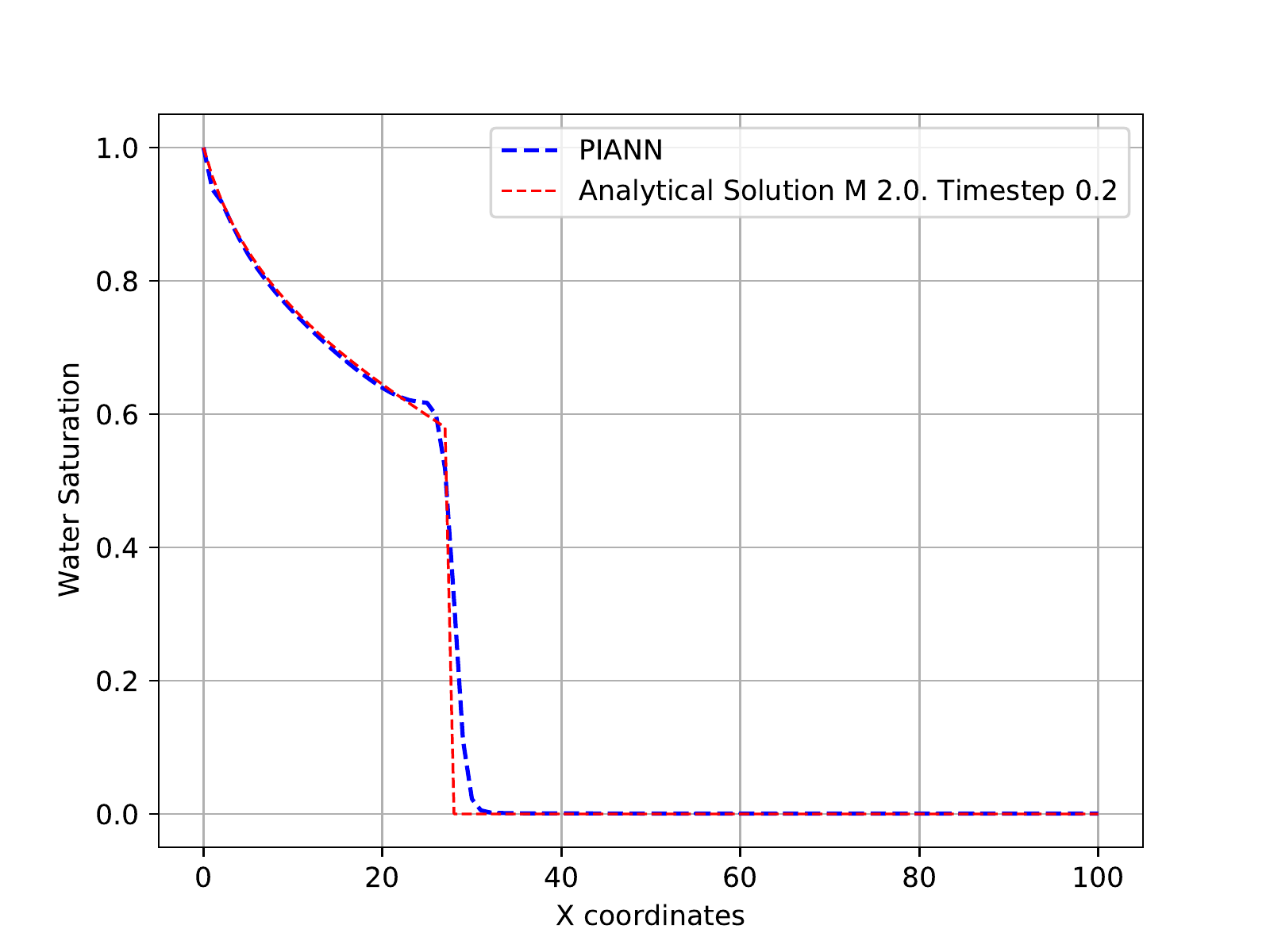}
     \includegraphics[width=0.28\linewidth]{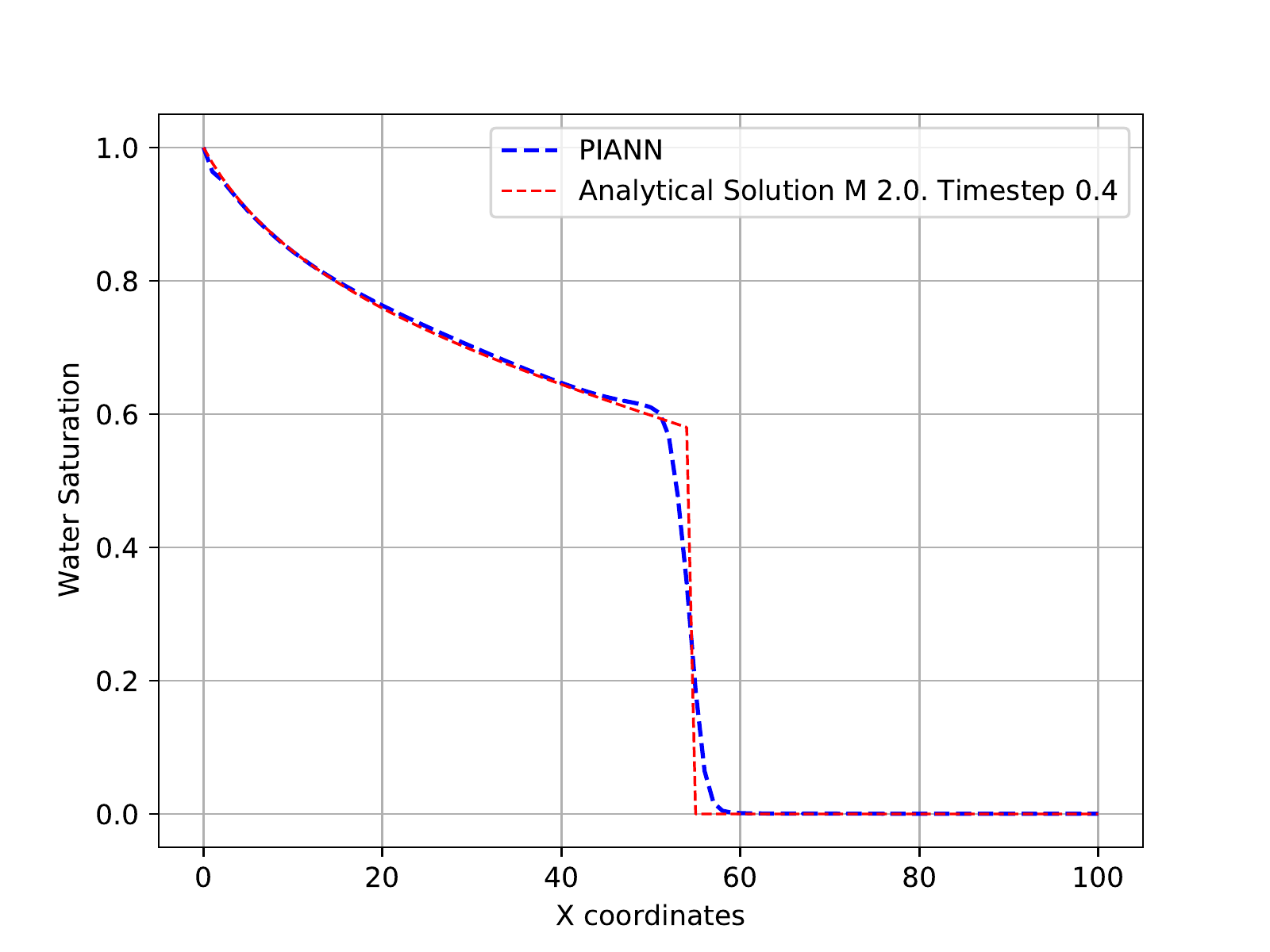}
     \includegraphics[width=0.28\linewidth]{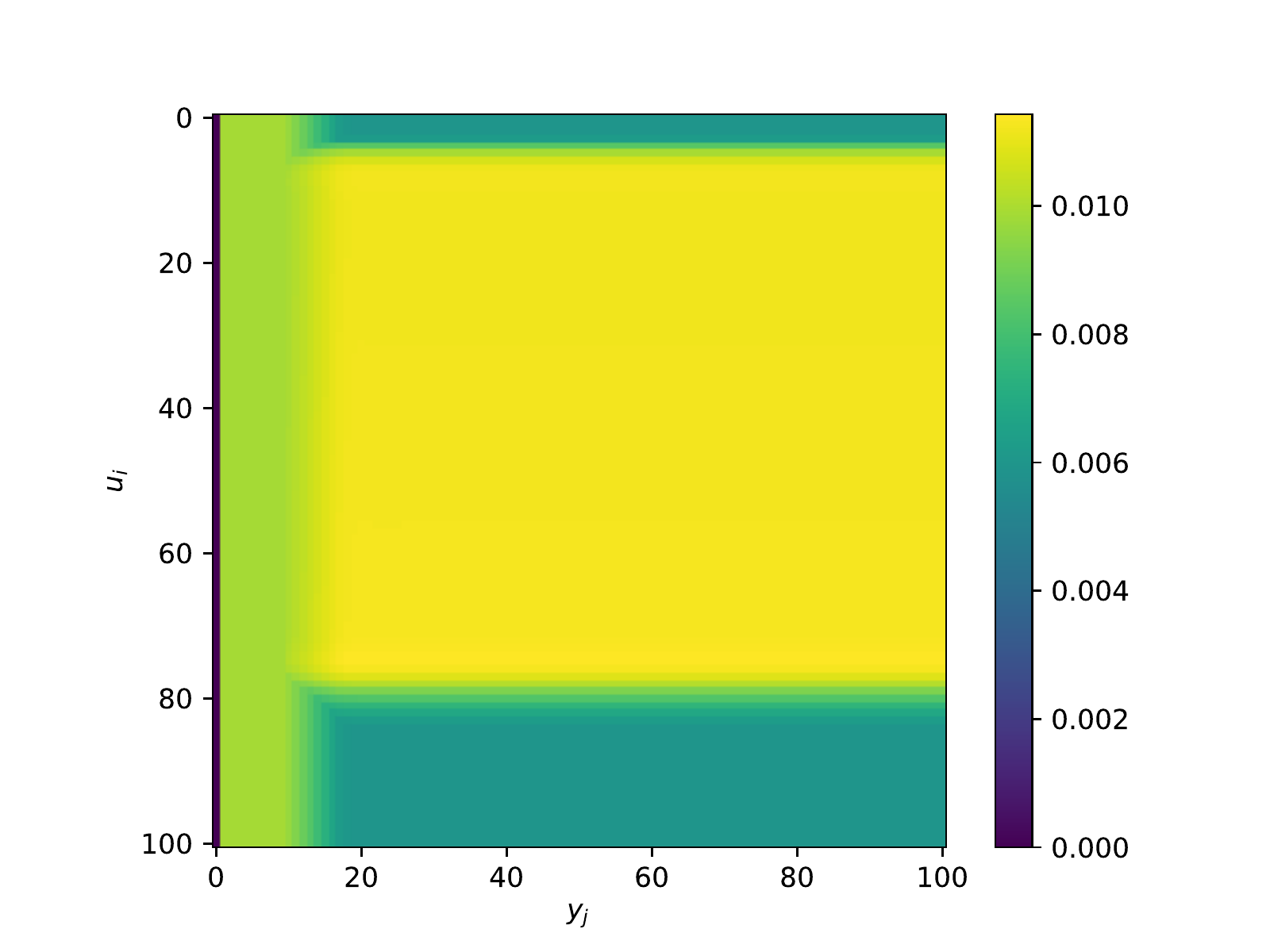}
     \includegraphics[width=0.28\linewidth]{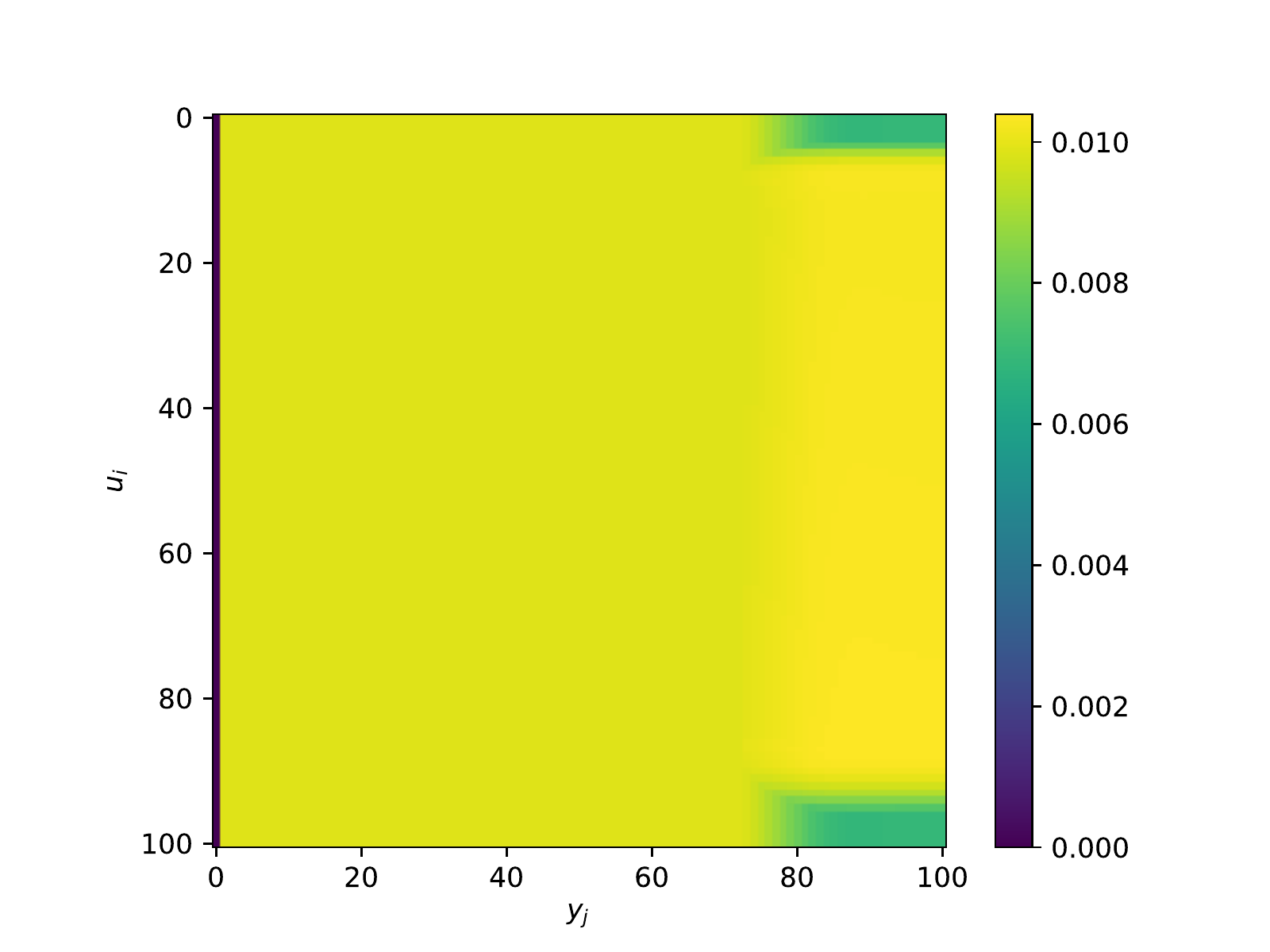}
     \includegraphics[width=0.28\linewidth]{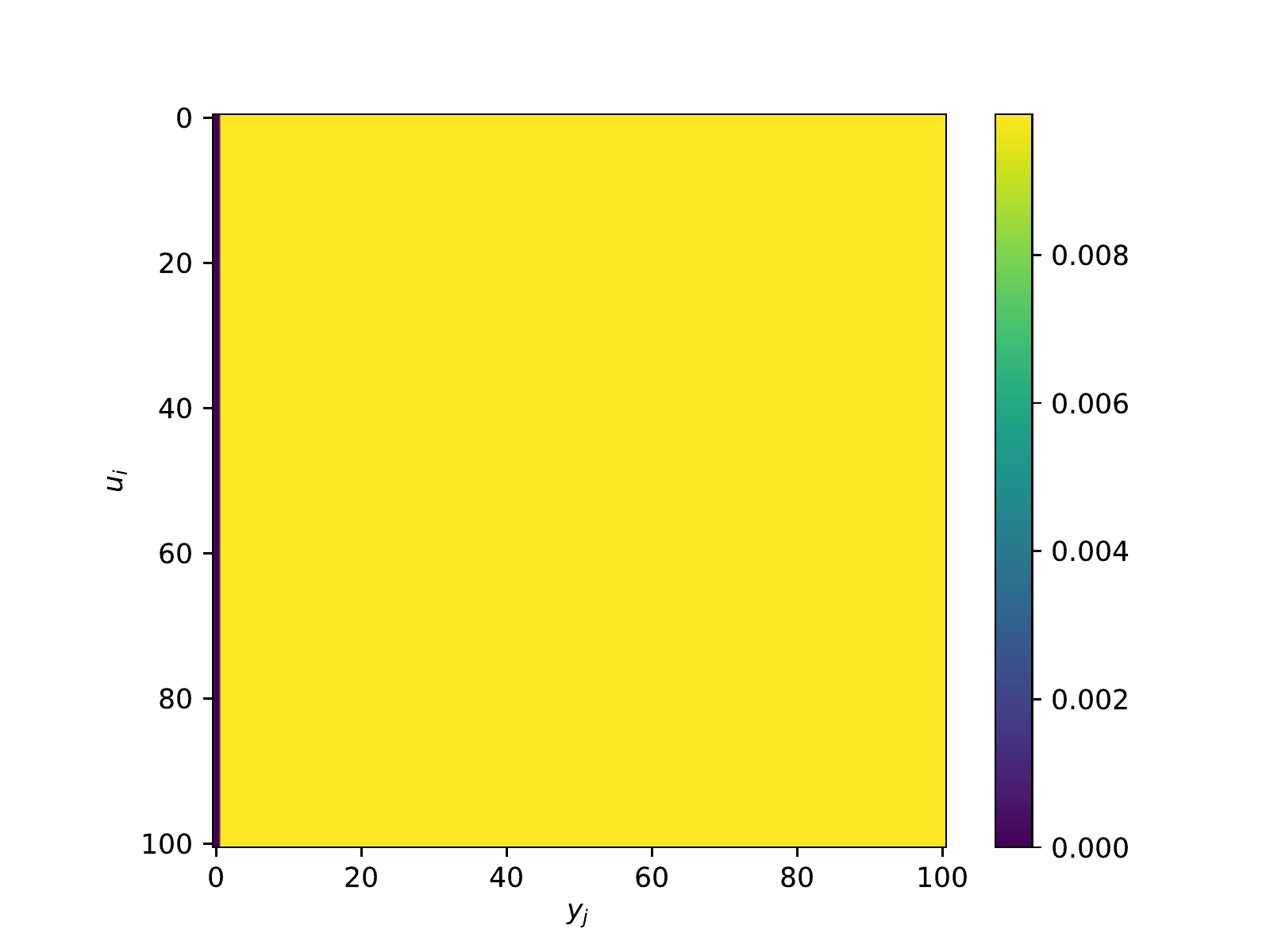}
     \includegraphics[width=0.28\linewidth]{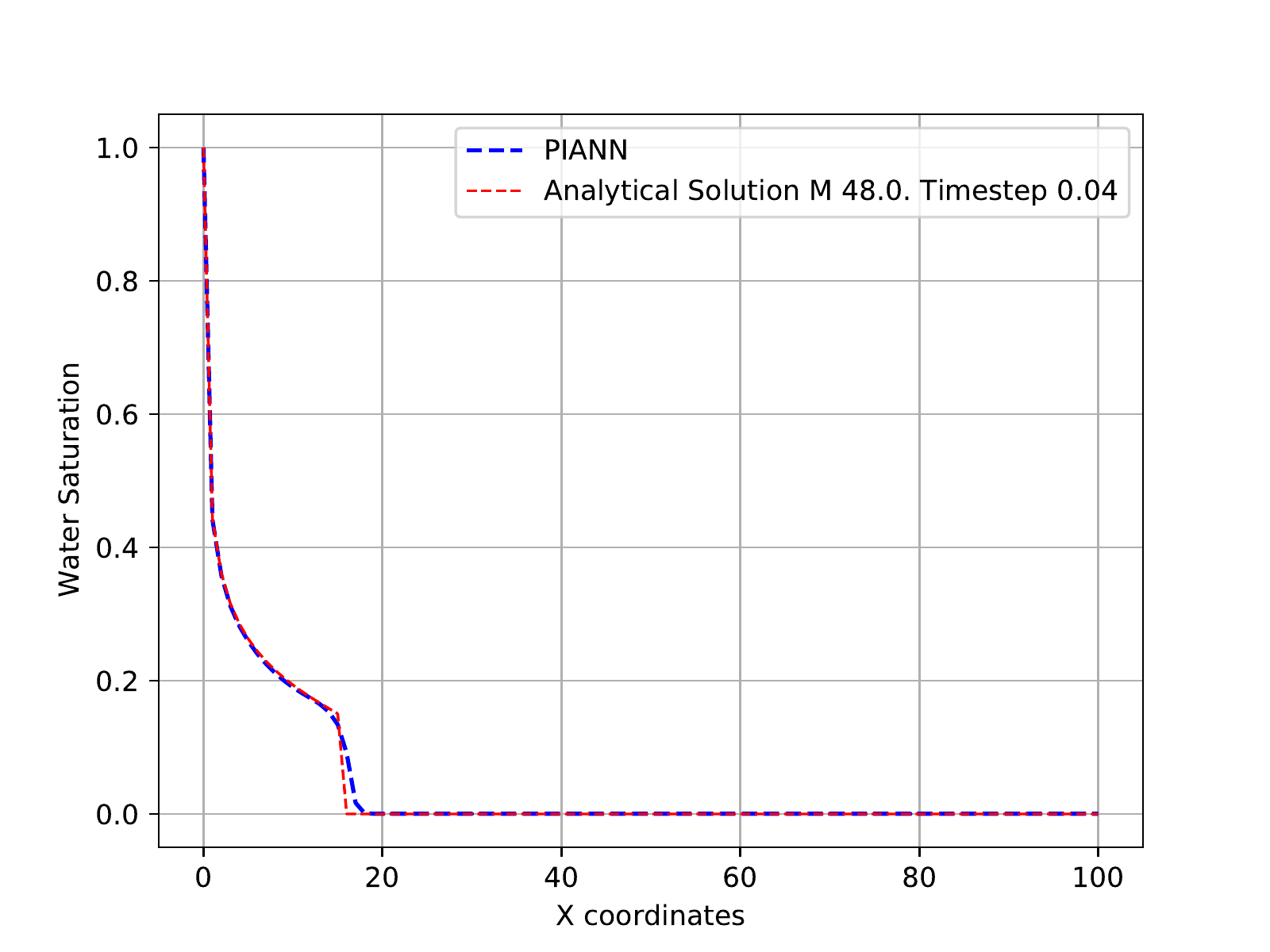}
     \includegraphics[width=0.28\linewidth]{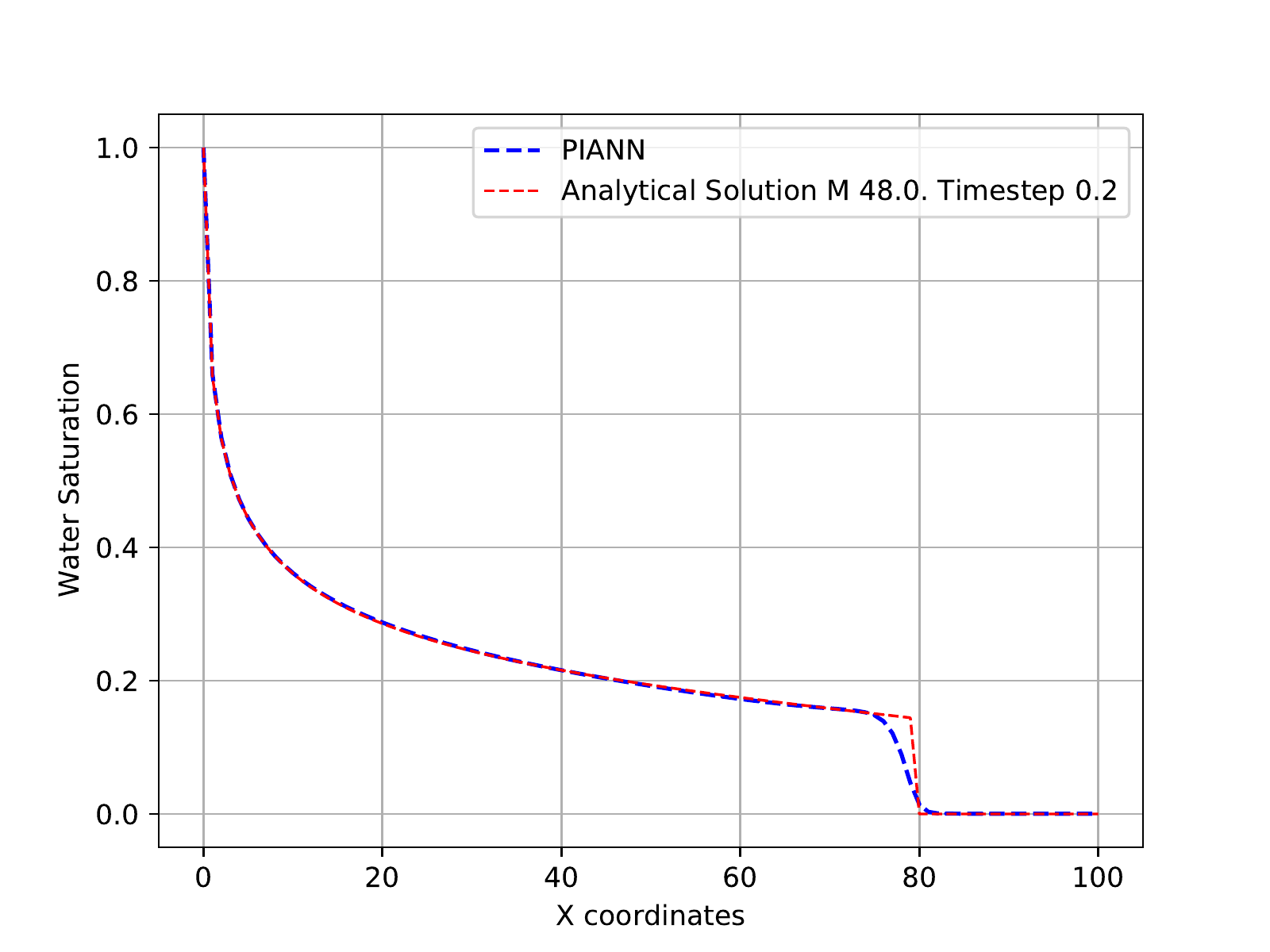}
     \includegraphics[width=0.28\linewidth]{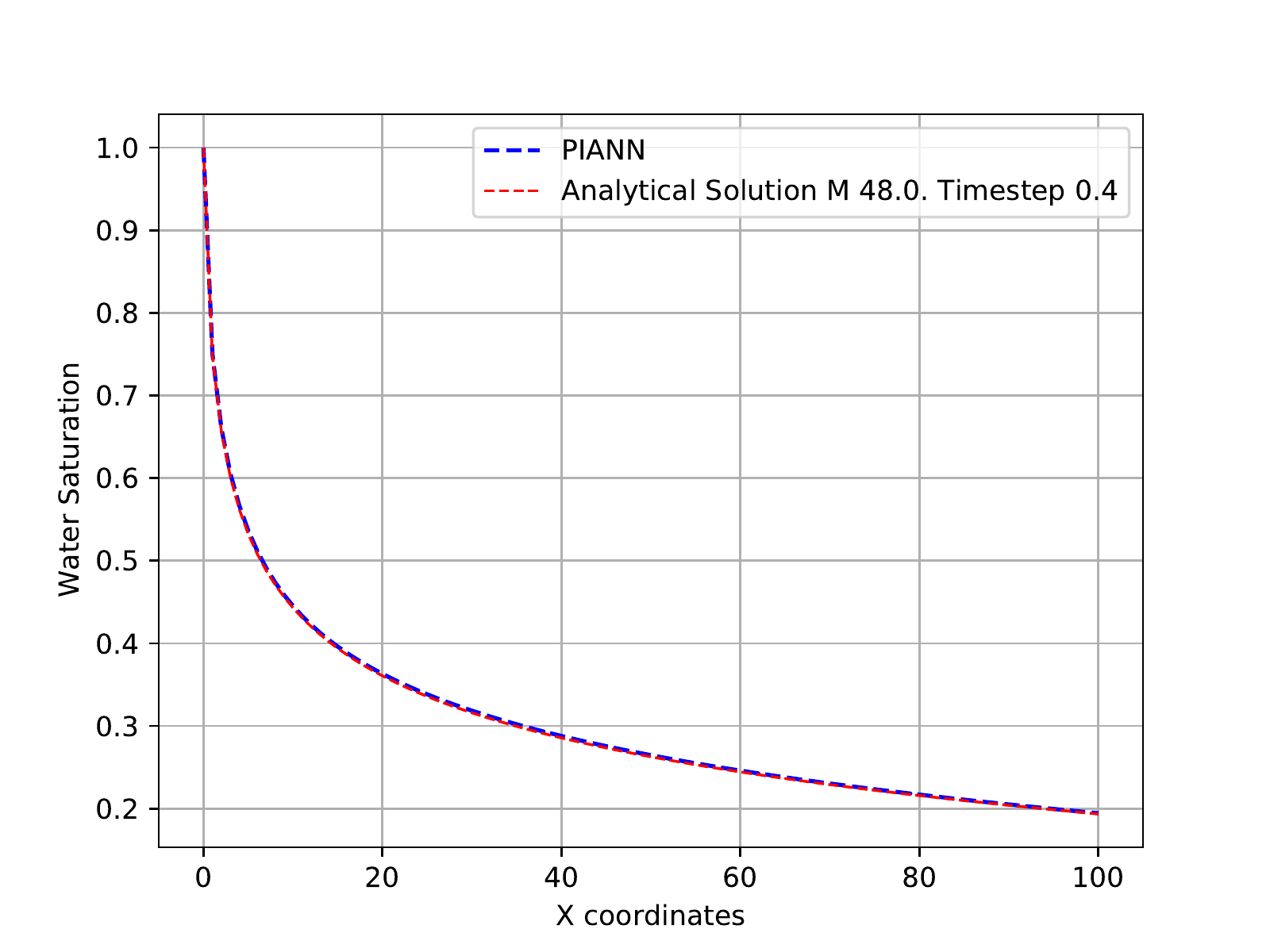}
     \includegraphics[width=0.28\linewidth]{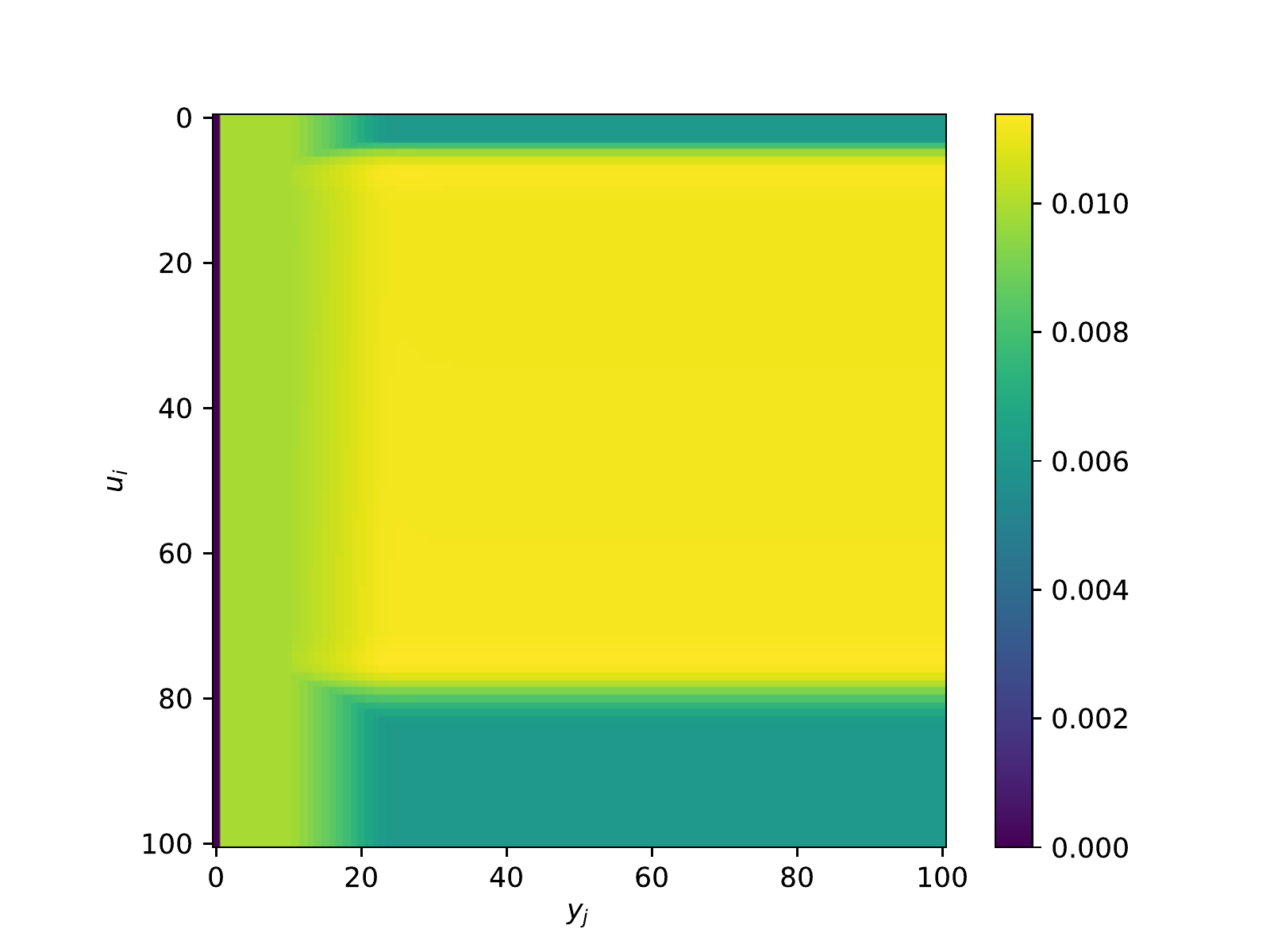}
     \includegraphics[width=0.28\linewidth]{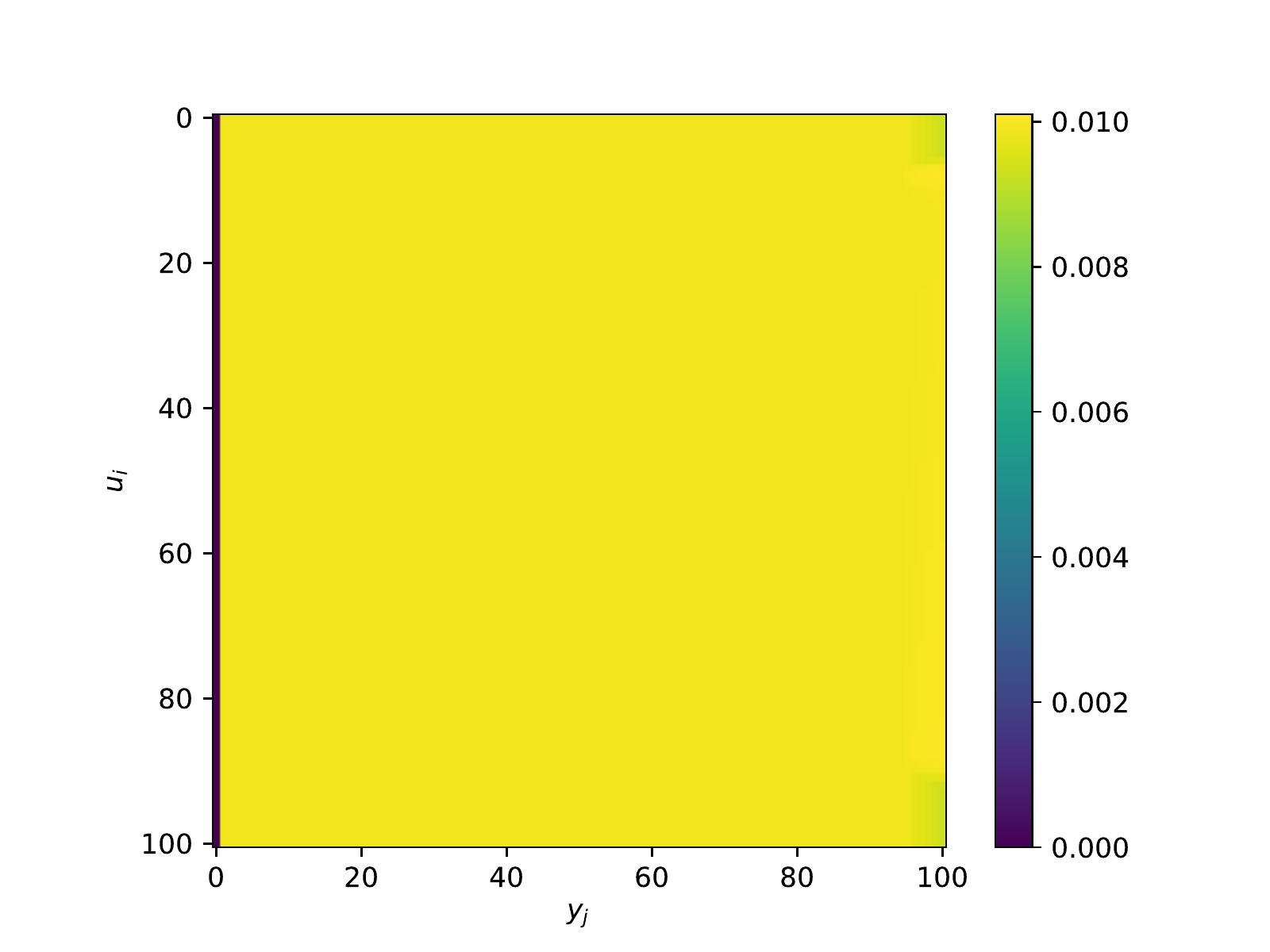}
     \includegraphics[width=0.28\linewidth]{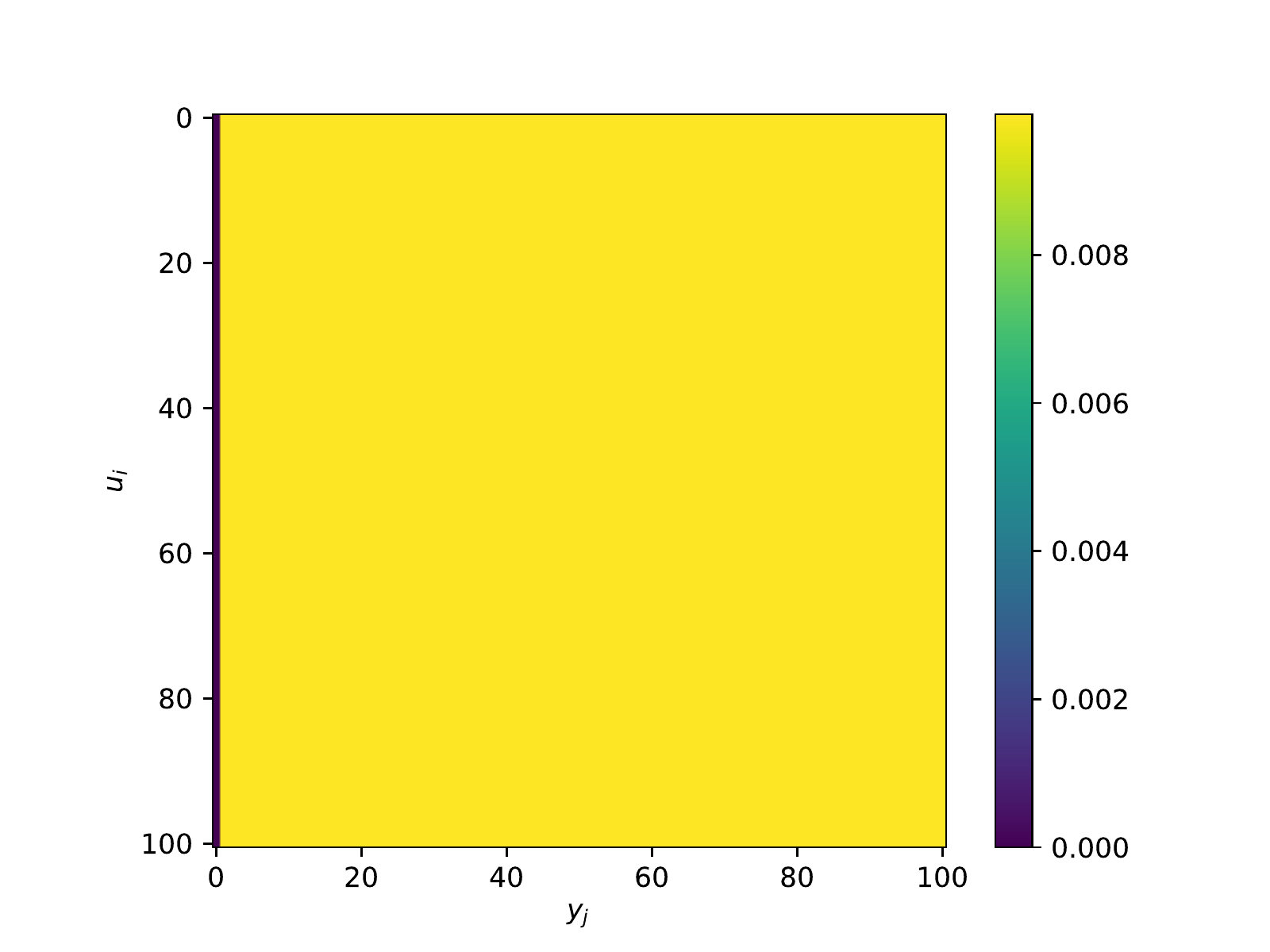}
     \includegraphics[width=0.28\linewidth]{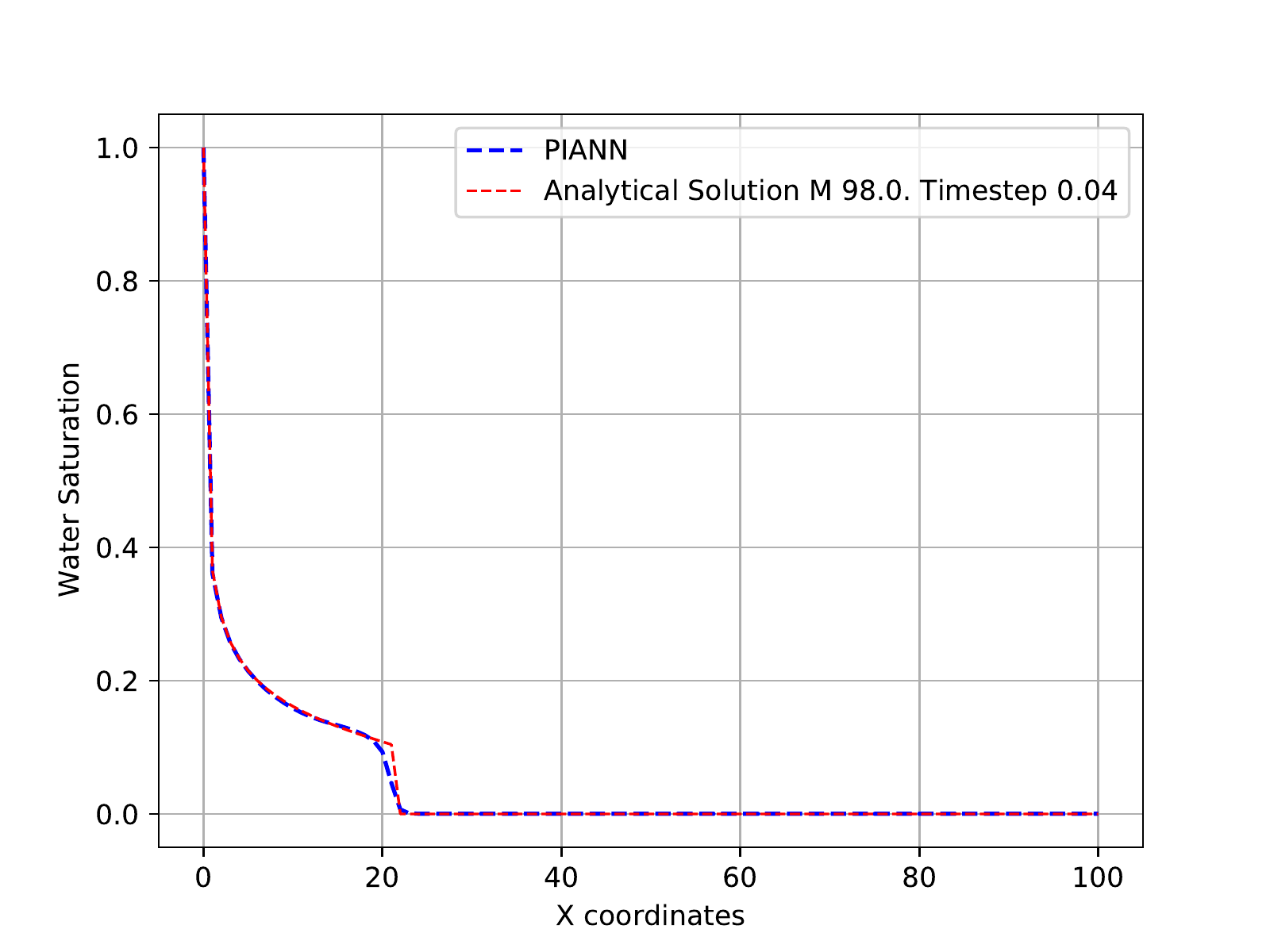}
     \includegraphics[width=0.28\linewidth]{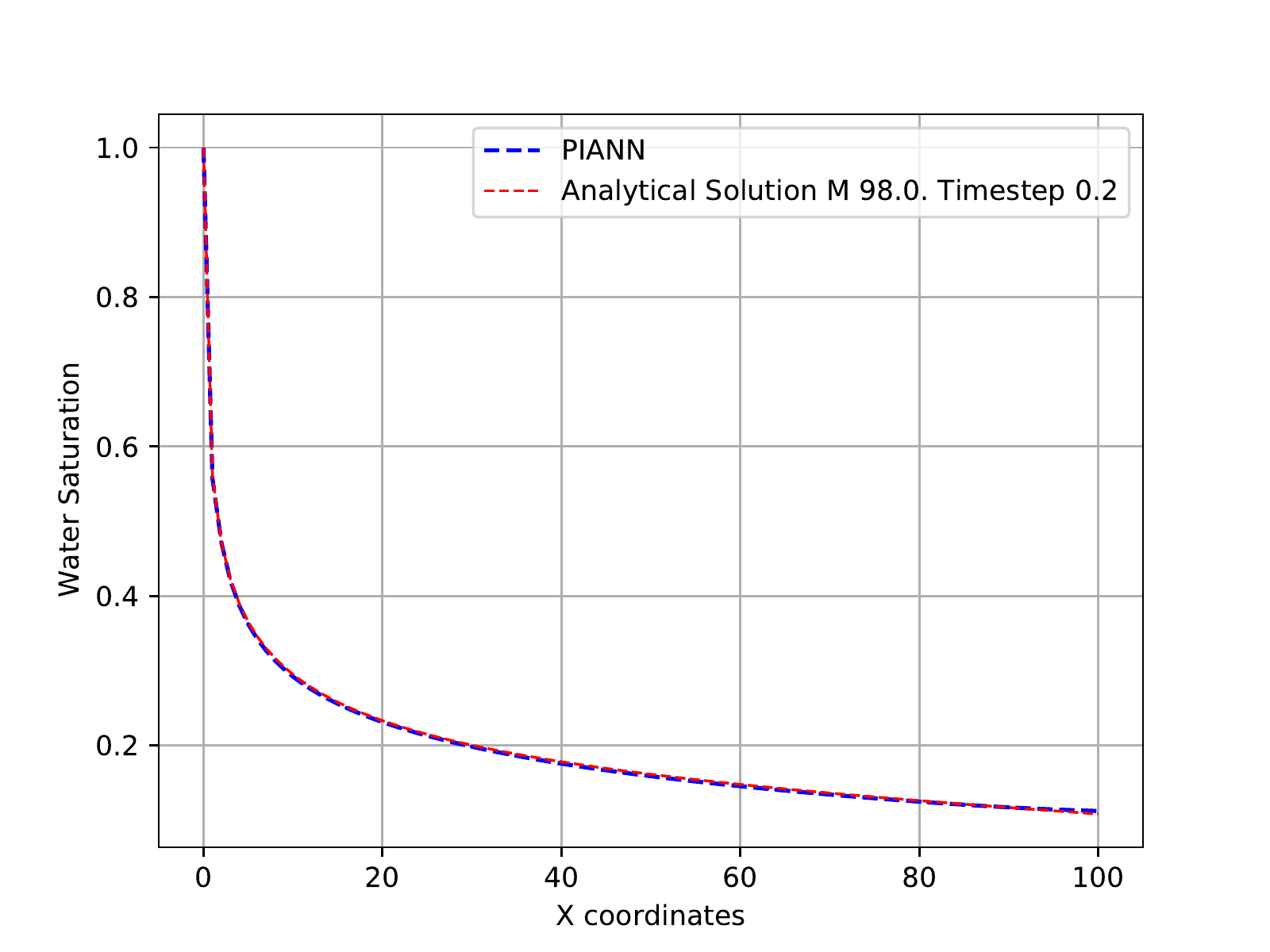}
     \includegraphics[width=0.28\linewidth]{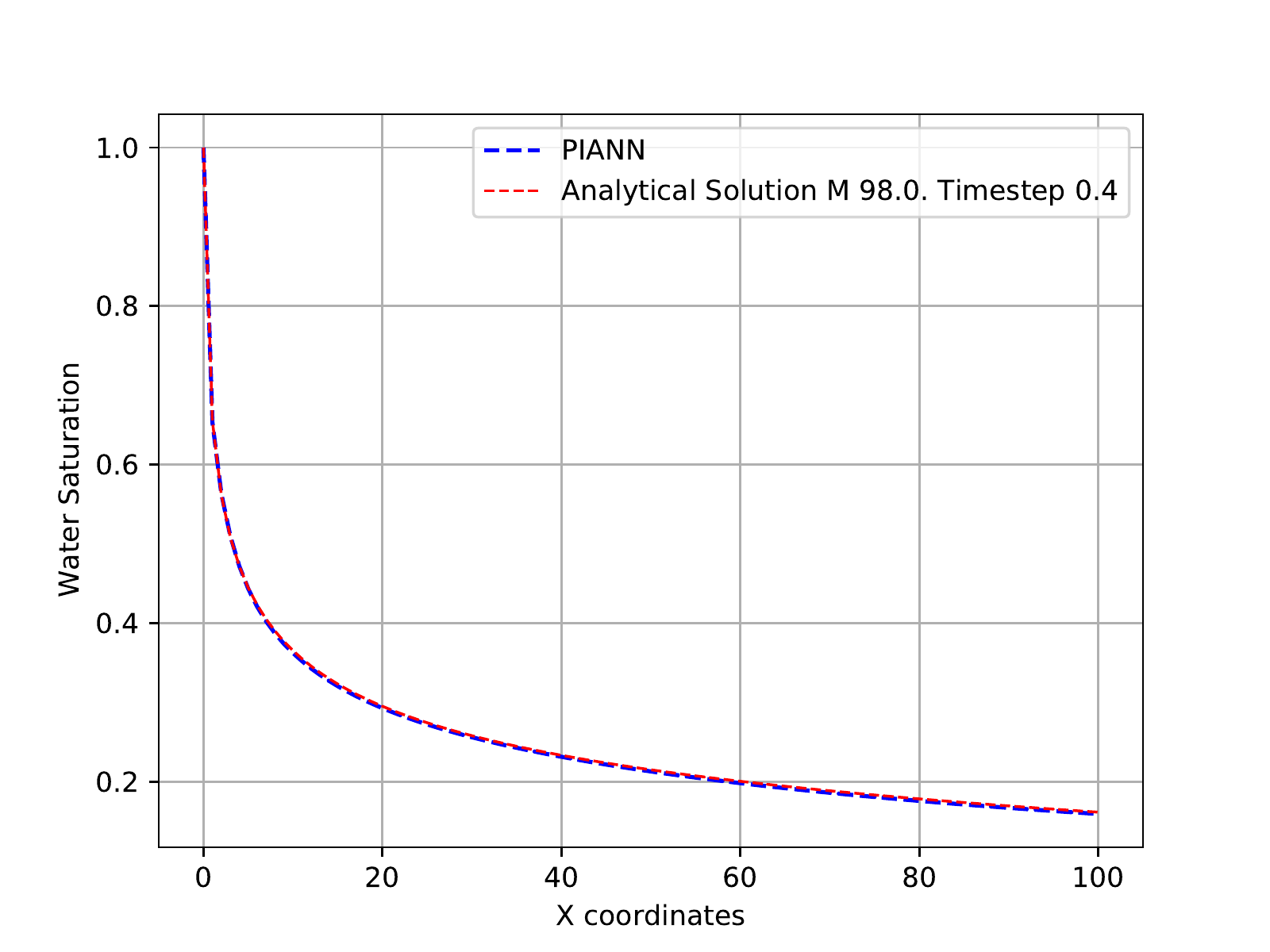}
     \caption{Top and bottom rows correspond to $ M = 2$ and $ M = 48 $ and $M=98$ for attention weights map and comparison of the predicted by the neural network and the exact solutions of the PDE, respectively. The columns from left to right, correspond to different time steps $t = 0.04$, $t = 0.20$ and $t=0.40$}
     \label{fig:solution}
 \end{figure*}

Figure \ref{fig:solution} shows the comparison between the analytical solution (red) and the solution obtained by our PIANN (blue) for different $M$ used during training. Top, middle and bottom rows correspond to $ M = 2$, $ M = 48 $ and $M=98$, respectively, and the columns from left to right, correspond to different time steps $t = 0.04$, $t = 0.20$, and $t = 0.40$, respectively. We can distinguish three regions of interest in the solution. Following increasing x coordinates, the first region on the left is one where the water saturation varies smoothly following the rarefaction part of the solution. The second region is a sharp saturation change that corresponds to the shock in the solution and the third region is ahead of the shock, with undisturbed water saturation values that are still at zero. For all cases, we observe that the PIANN properly learns the correct rarefaction behavior of the first region and approximates the analytical solution extremely well. In the third region, the PIANN also fits to the analytical solution perfectly and displays an undisturbed water saturation at zero. As for any classical numerical methods, the shock region is the most challenging to resolve.

Around the shock, the PIANN seems unable to perfectly resolve a sharp front, and the represented behavior is a shock that is smoothed over and displays non-monotonic artifacts upstream and downstream of the front. The location of the shock is, however, well captured. Such a behavior is reminiscent of the behavior observed in higher-order finite difference methods, where slope-limiters are often used to correct for the non-physical oscillations. 

\begin{figure}
\centering
\includegraphics[width=.5\linewidth]{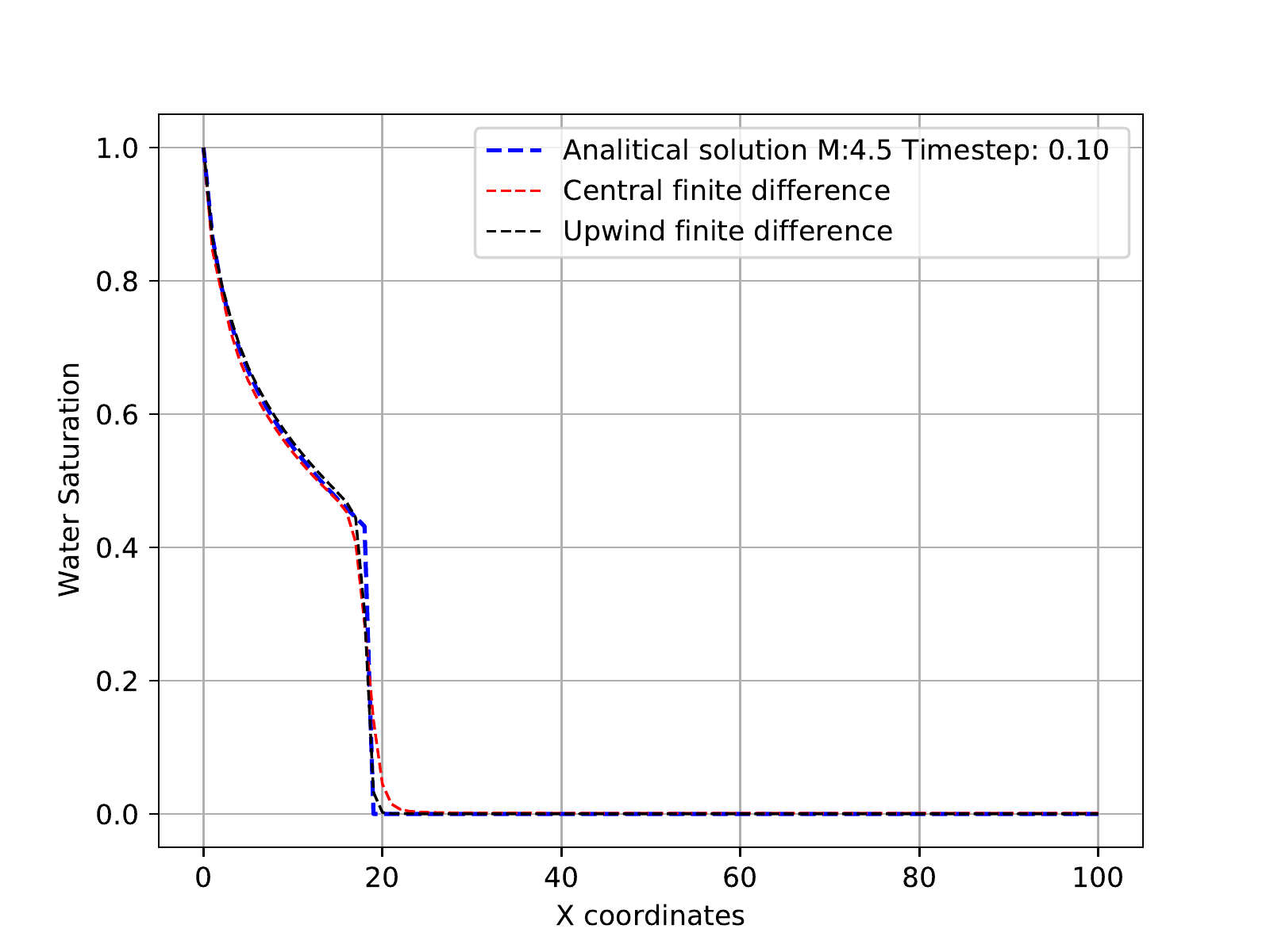}
\caption{Comparison between solutions obtained with residuals evaluated using central and upwind finite differences, for $M=4.5$.}
\label{fig:residuals_comparation}
\end{figure}

Importantly, it means the PIANN needs to learn where the shock is located in order to fit differently to both sides of it. This is the role of the attention mechanism of our architecture. On top of each figure we have visualized the attention map introduced by~\cite{bahdanau2014neural} for every timestep. These maps visualize the attention weight $\balpha_{i,j}$ to predict the variable $\bu_i$. We observe that in all cases the attention mechanism identifies the discontinuity, water front, and subsequently modifies the behavior of the network in the three different regions described above. This shows that attention mechanism provides all the necessary information to capture discontinuities automatically without the necessity of training data or a prior knowledge. Finally, it is important to note that attention weights of the attention mechanism are constant when the shock/discontinuity disappears. 

\begin{figure*}
     \centering
     \includegraphics[width=0.28\linewidth]{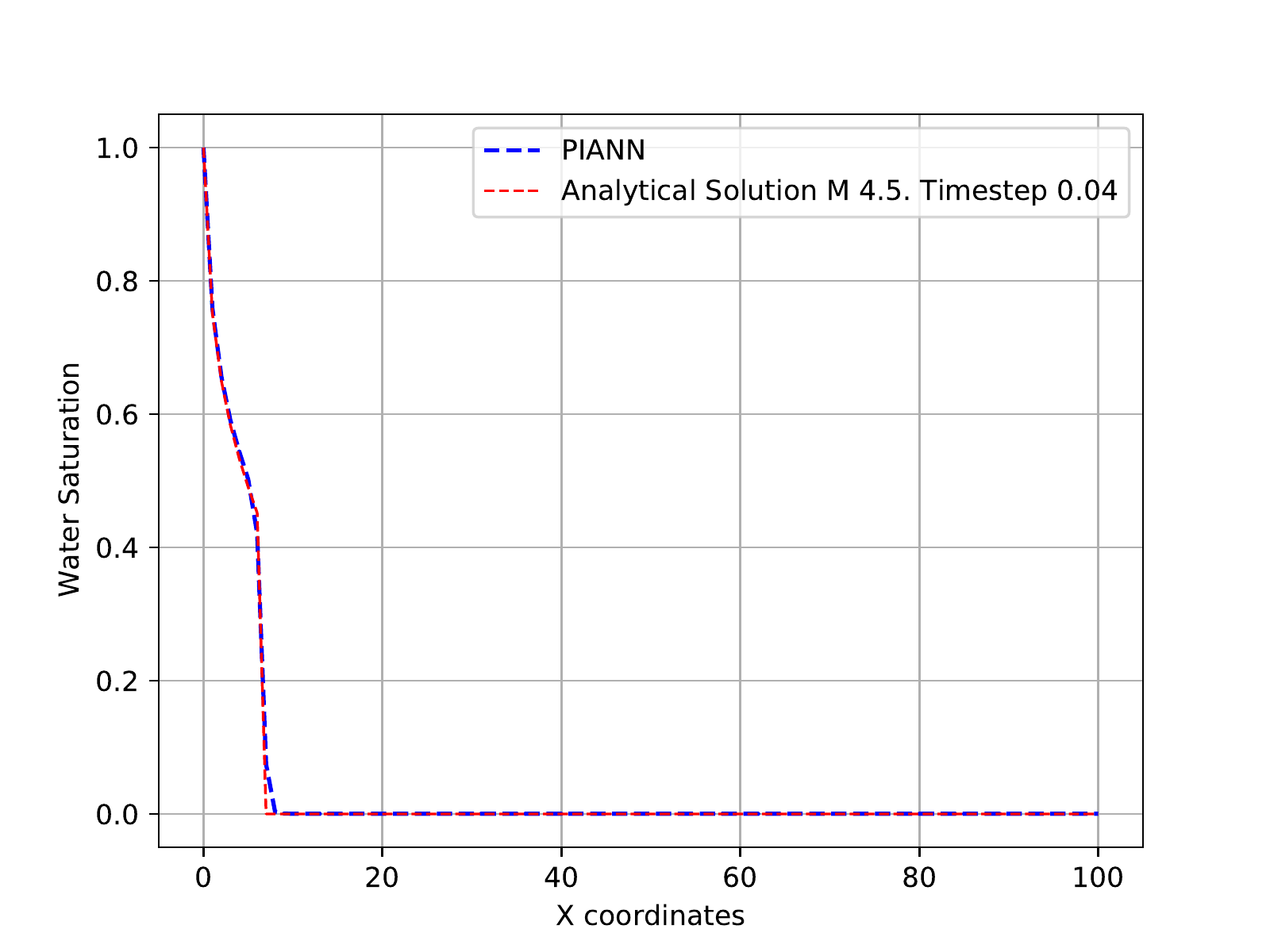}
     \includegraphics[width=0.28\linewidth]{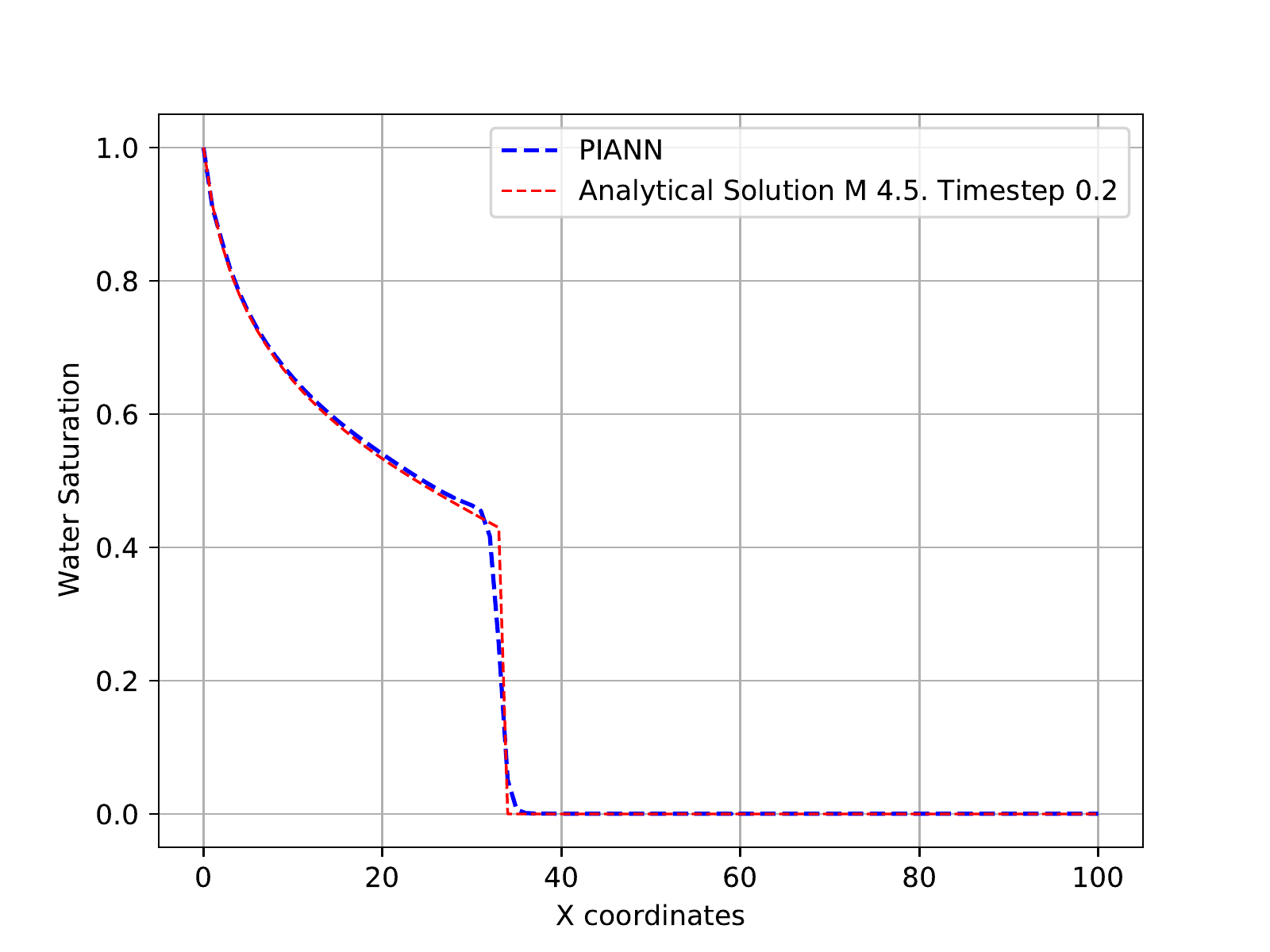}
     \includegraphics[width=0.28\linewidth]{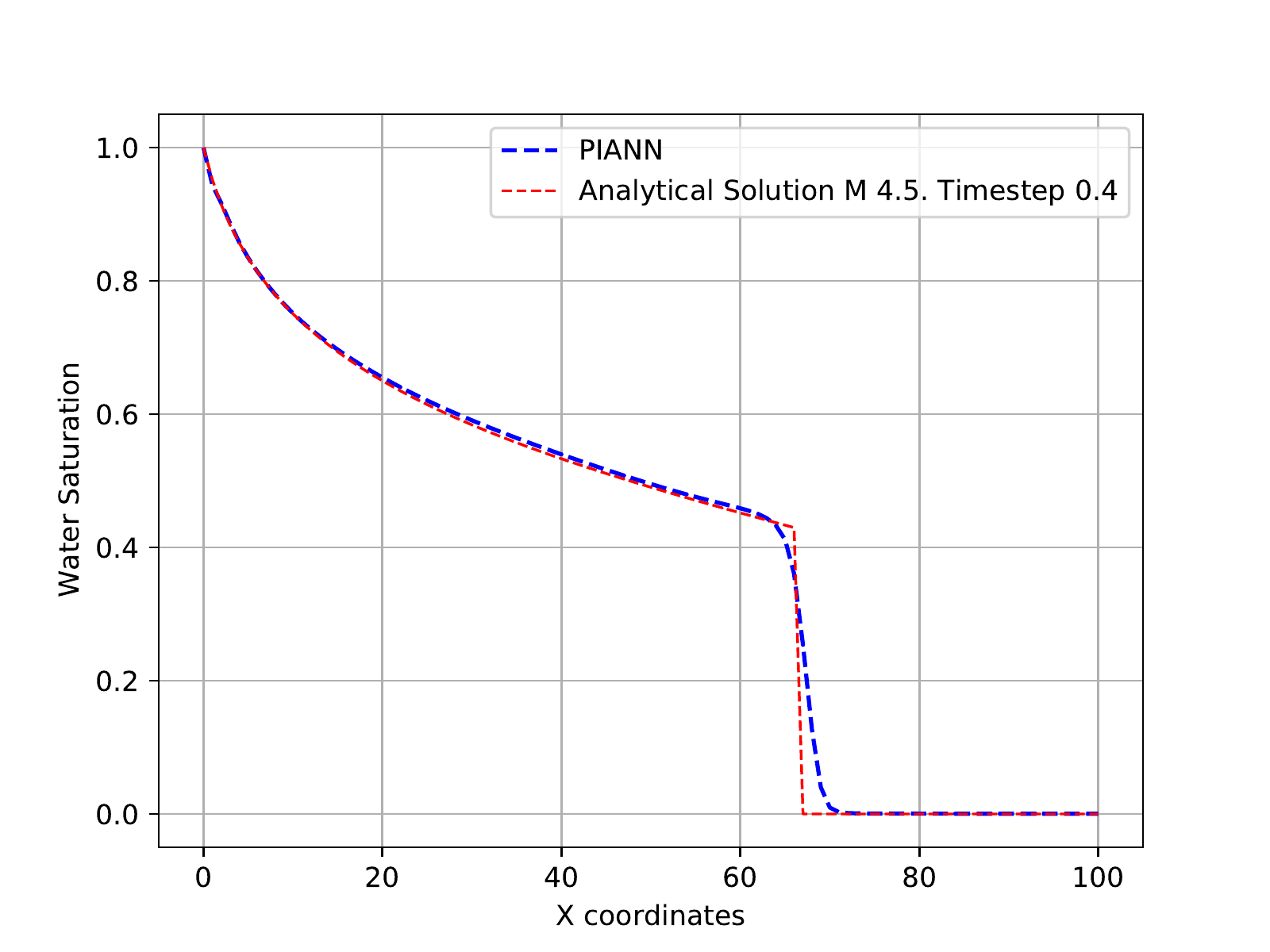}
     \includegraphics[width=0.28\linewidth]{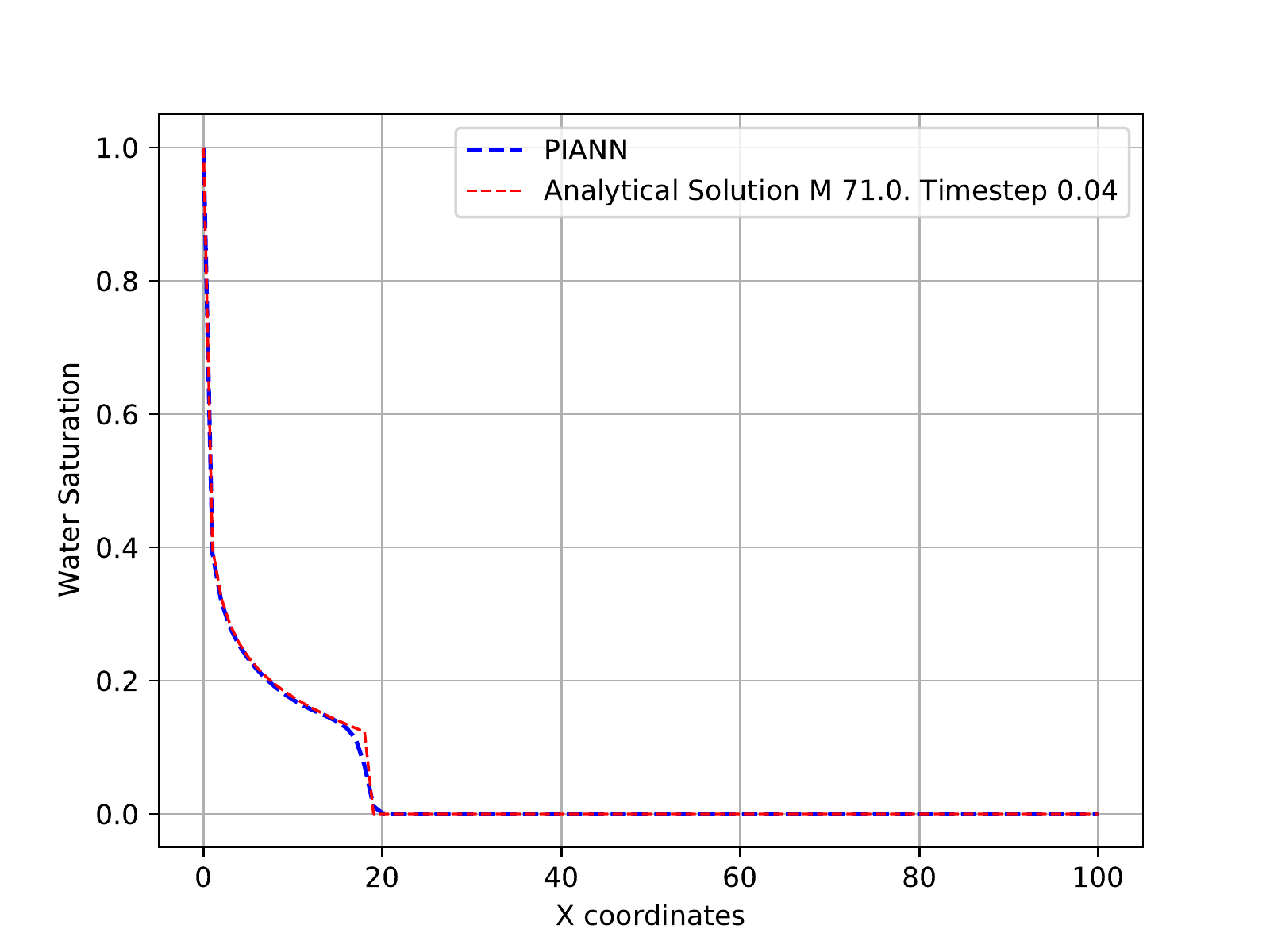}
     \includegraphics[width=0.28\linewidth]{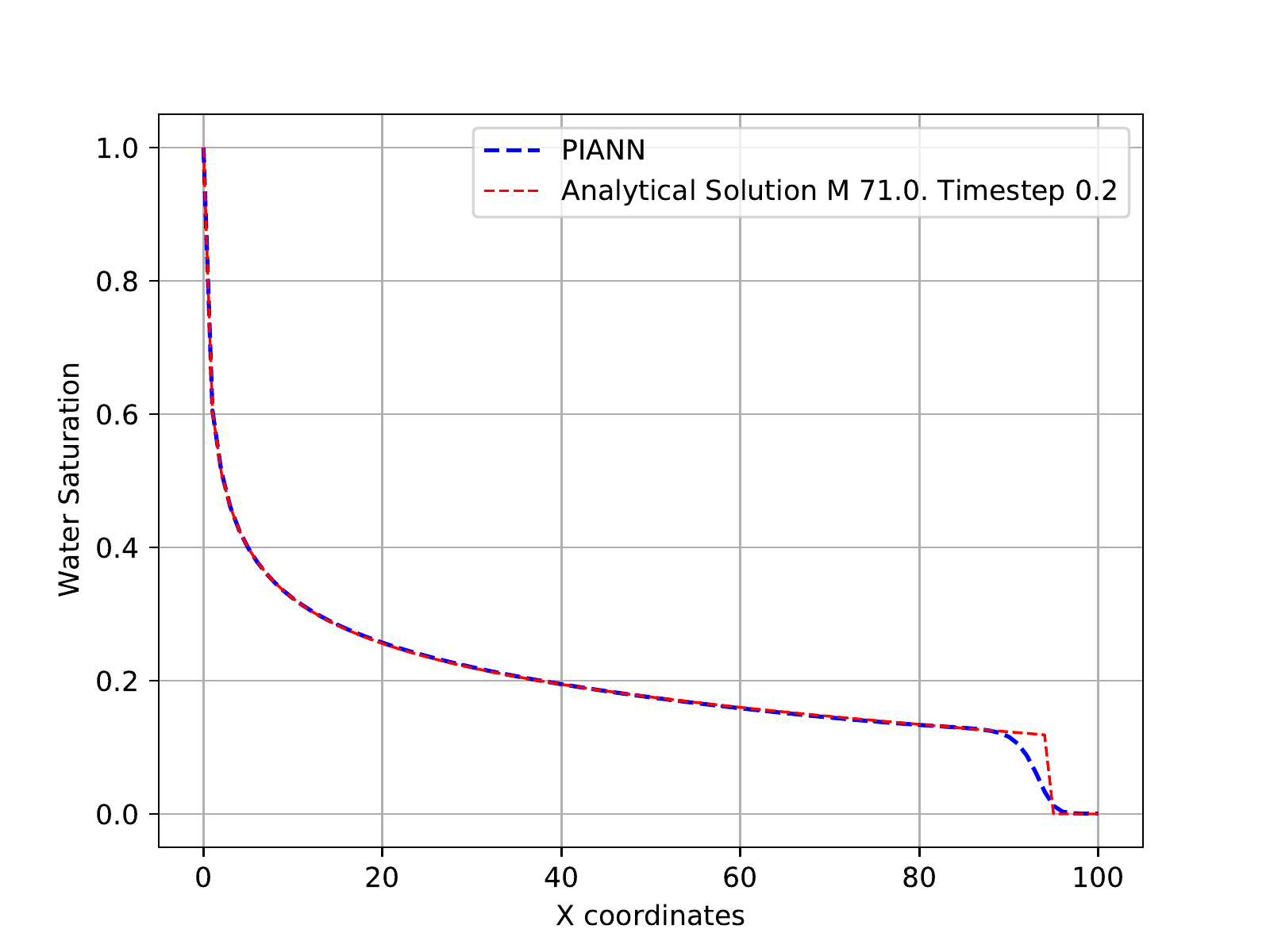}
     \includegraphics[width=0.28\linewidth]{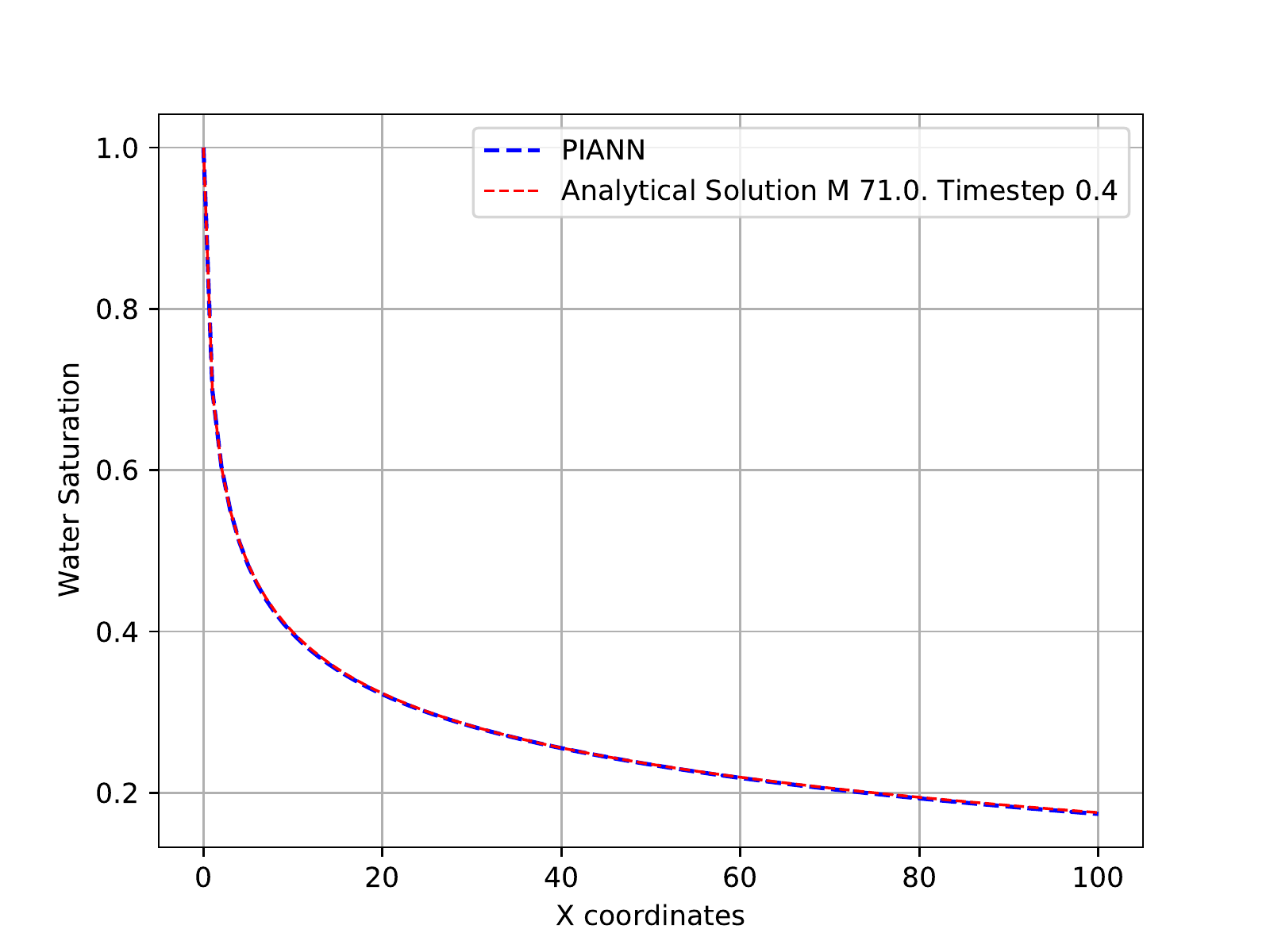}
     \caption{Top and bottom rows correspond to $ M = 4.5$ and $ M = 71 $ for comparison of the predicted by the neural network and the exact solutions of the PDE, respectively. The columns from left to right, correspond to different time steps $t = 0.04$, $t = 0.20$ and $t=0.40$}
     \label{fig:interpolation}
 \end{figure*}

In addition, we test the behavior of our methodology to provide solutions for BL at points within the training range of $M$: $M=4.5$ and $M=71$. In other words, we want to check the capability of our PIANN model to interpolate solutions. Figure \ref{fig:interpolation} shows that our PIANNs provide solutions that correctly detect the shock. 

\begin{figure*}
     \centering
     \includegraphics[width=0.28\linewidth]{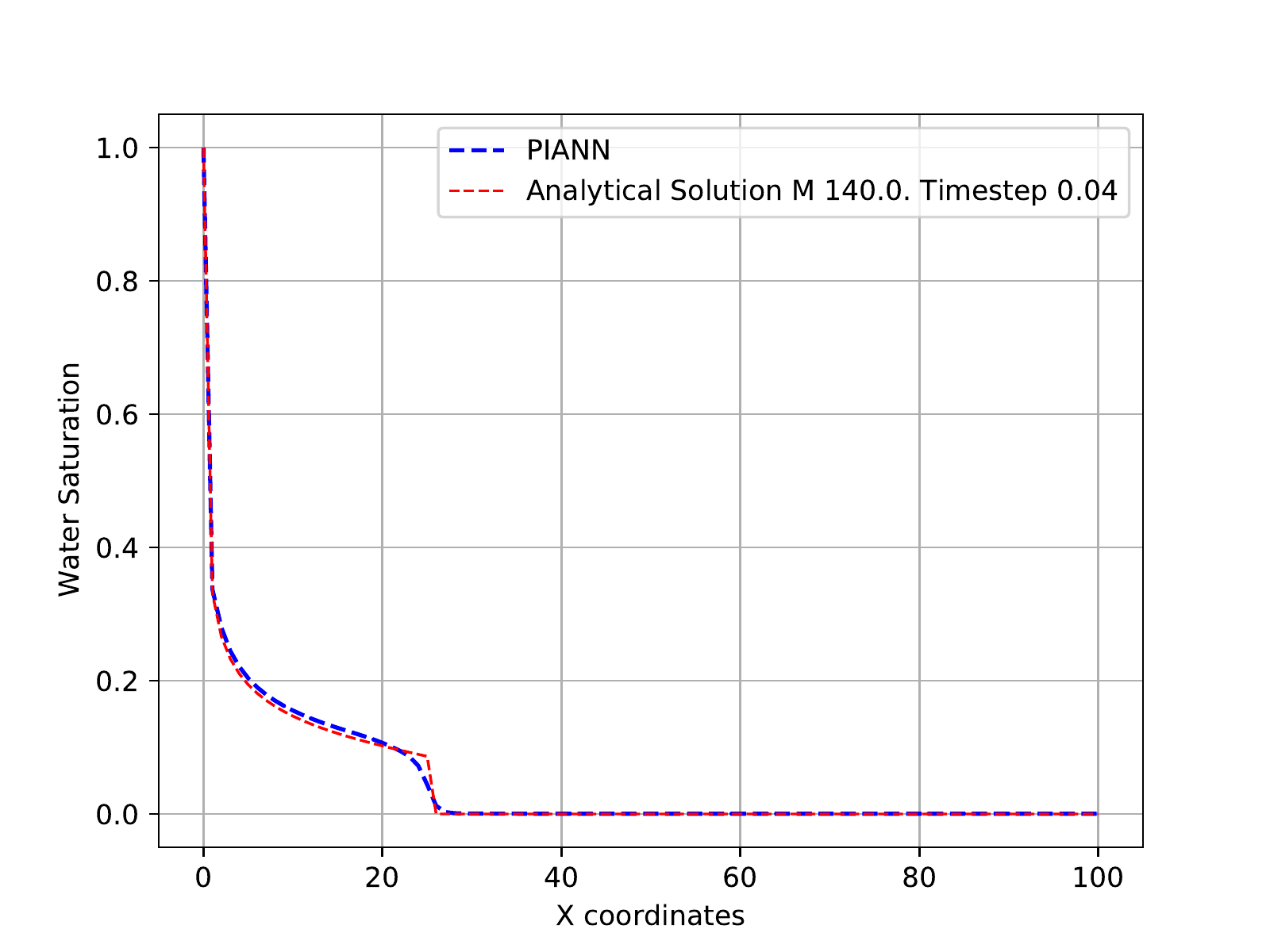}
     \includegraphics[width=0.28\linewidth]{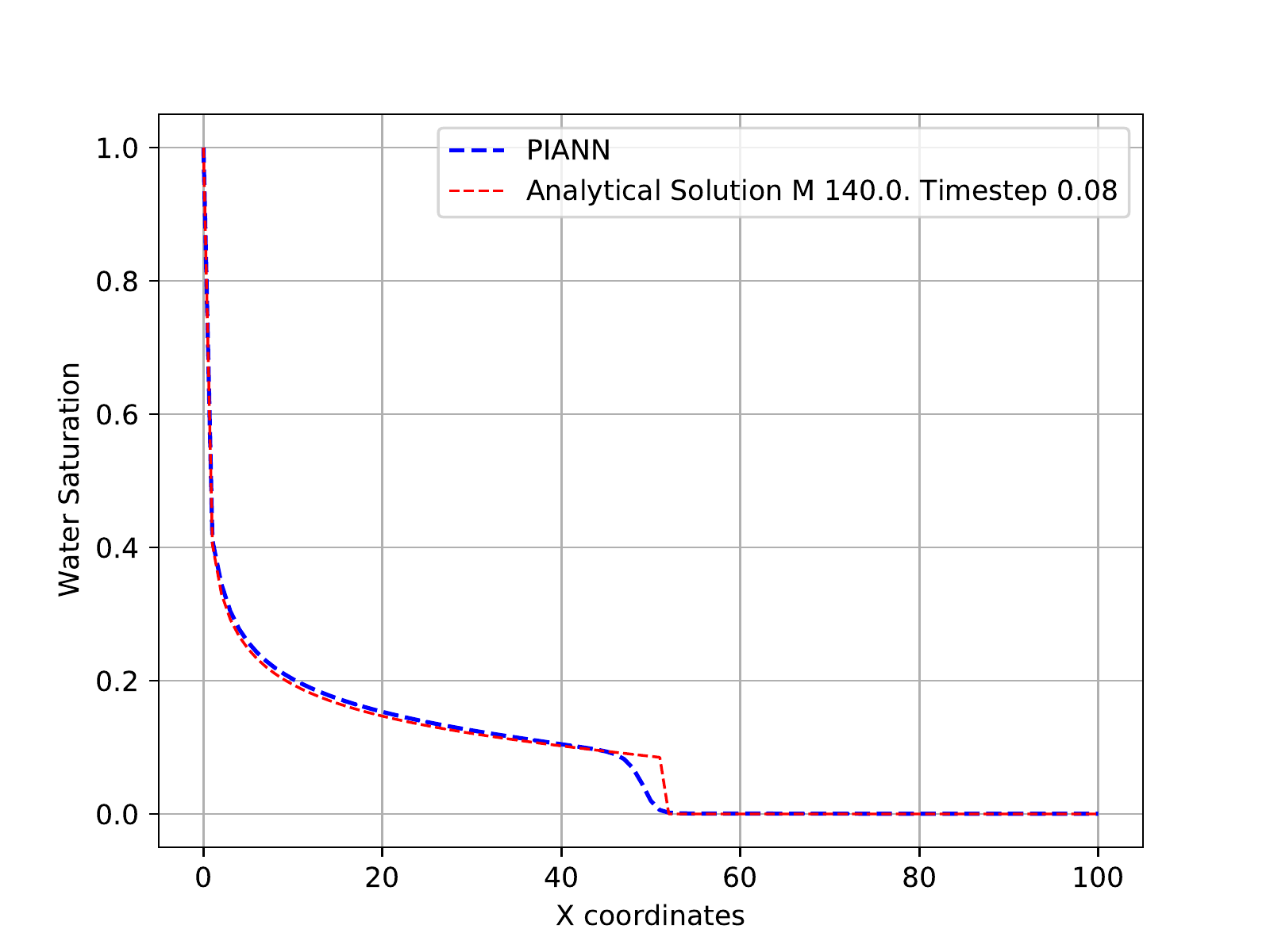}
     \includegraphics[width=0.28\linewidth]{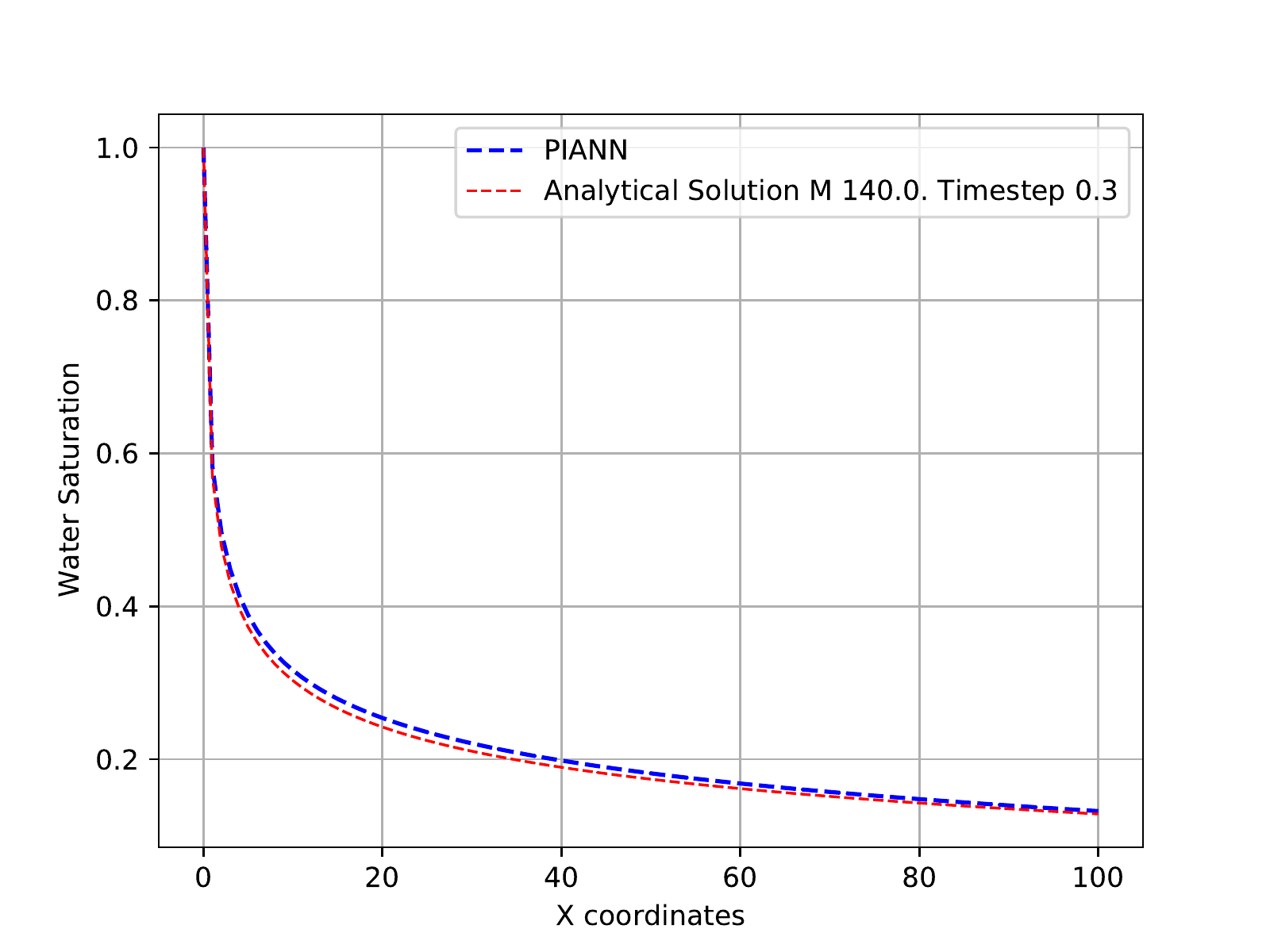}
     \includegraphics[width=0.28\linewidth]{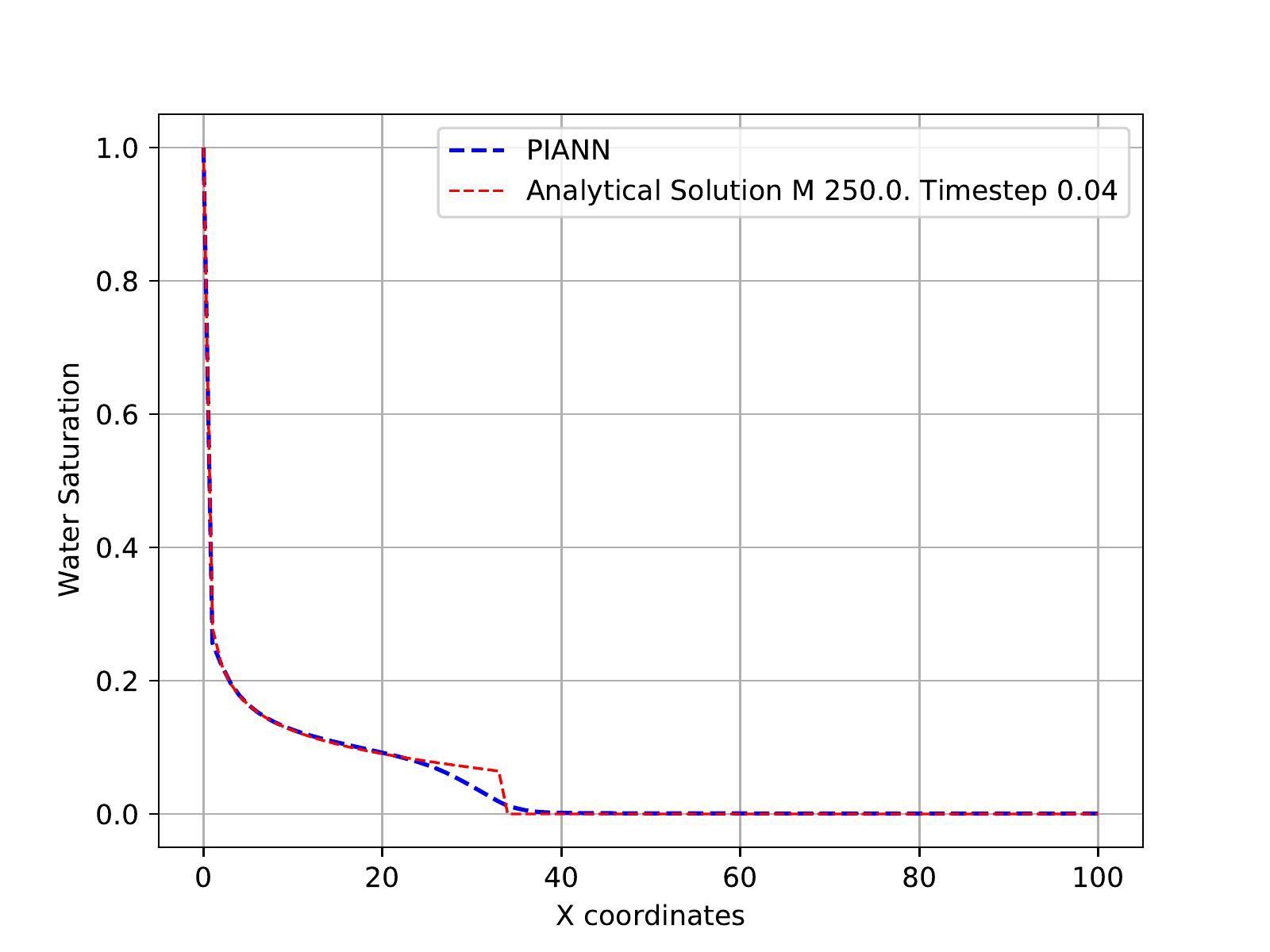}
     \includegraphics[width=0.28\linewidth]{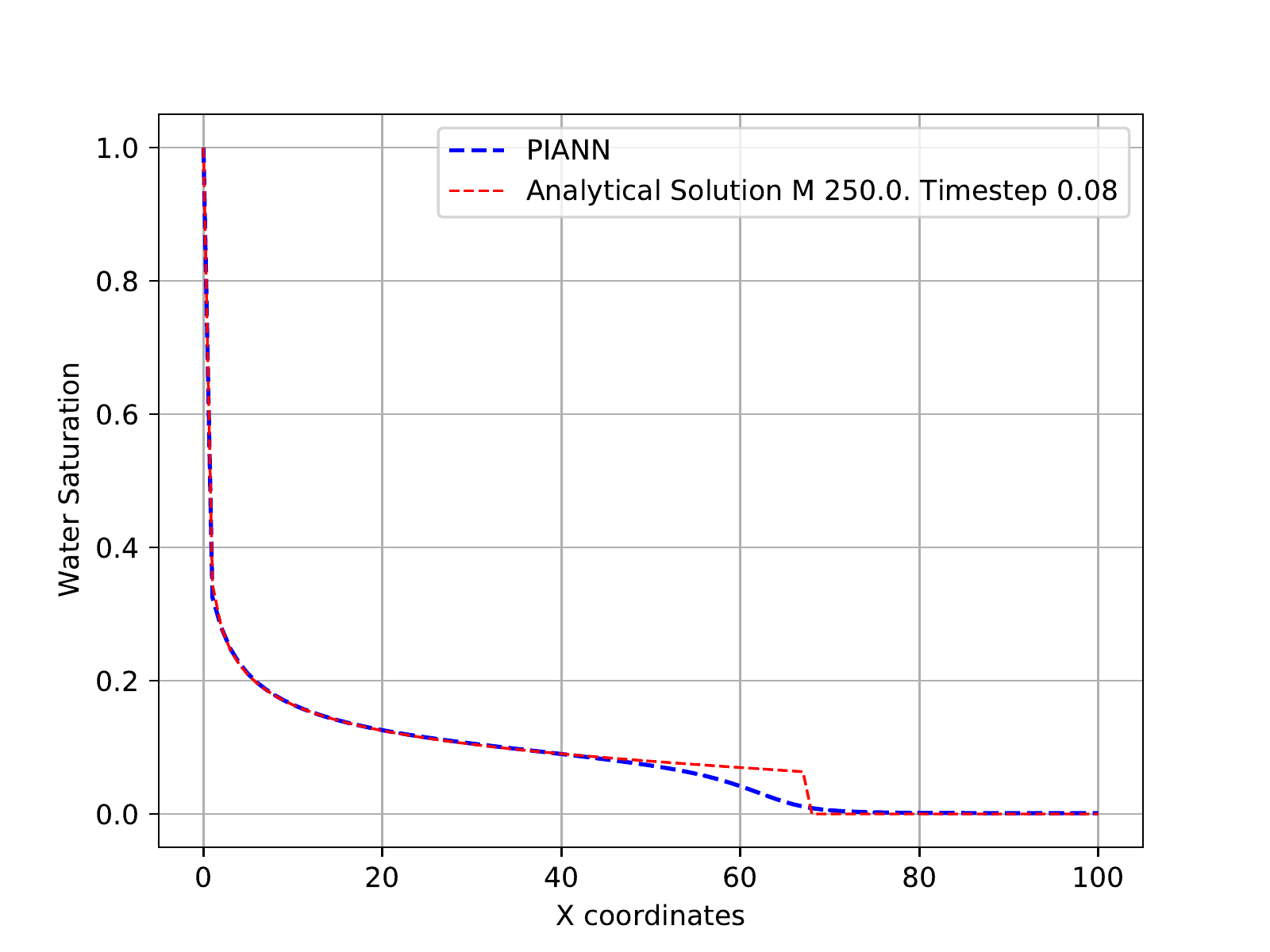}
     \includegraphics[width=0.28\linewidth]{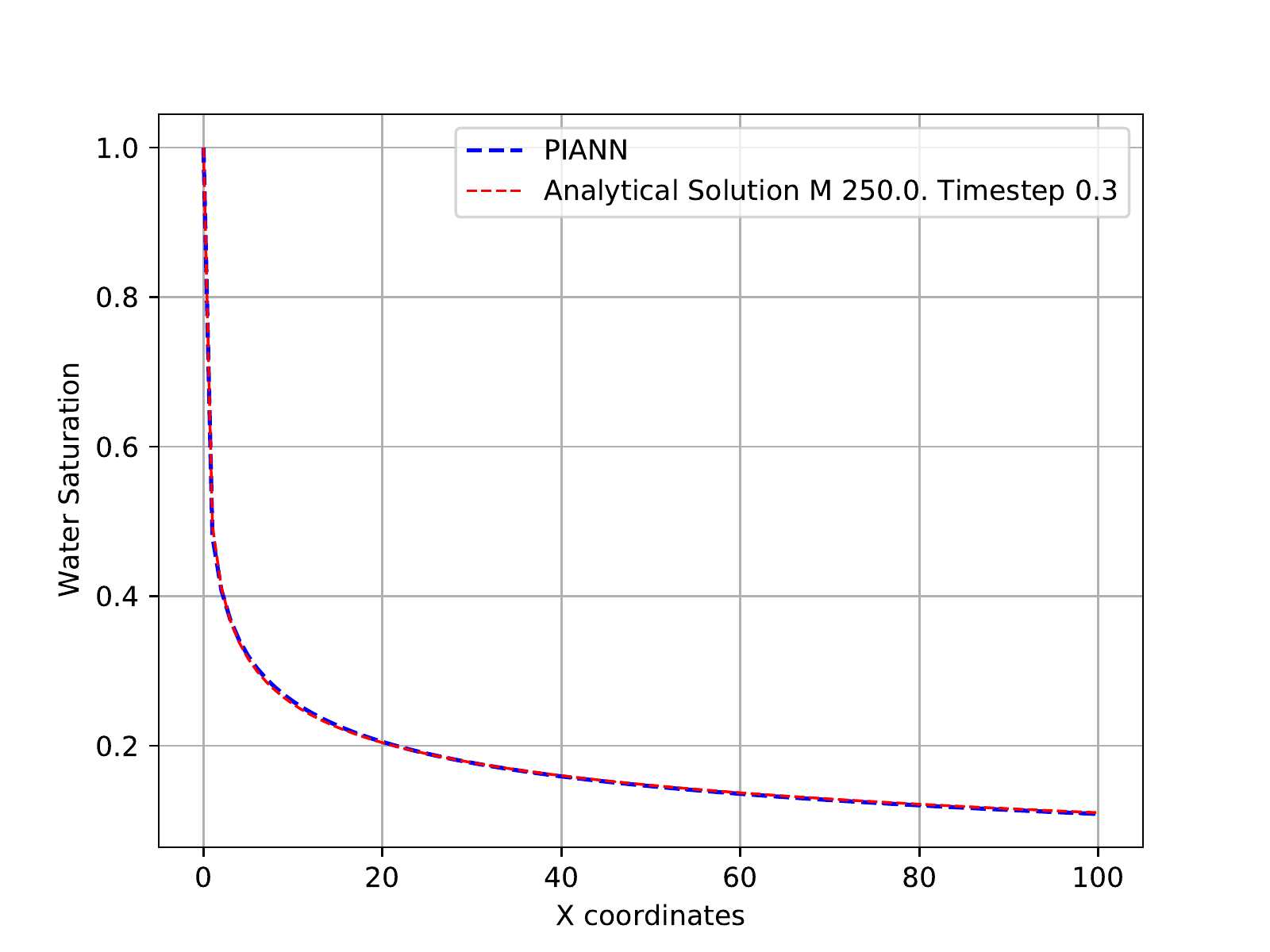}
     \includegraphics[width=0.28\linewidth]{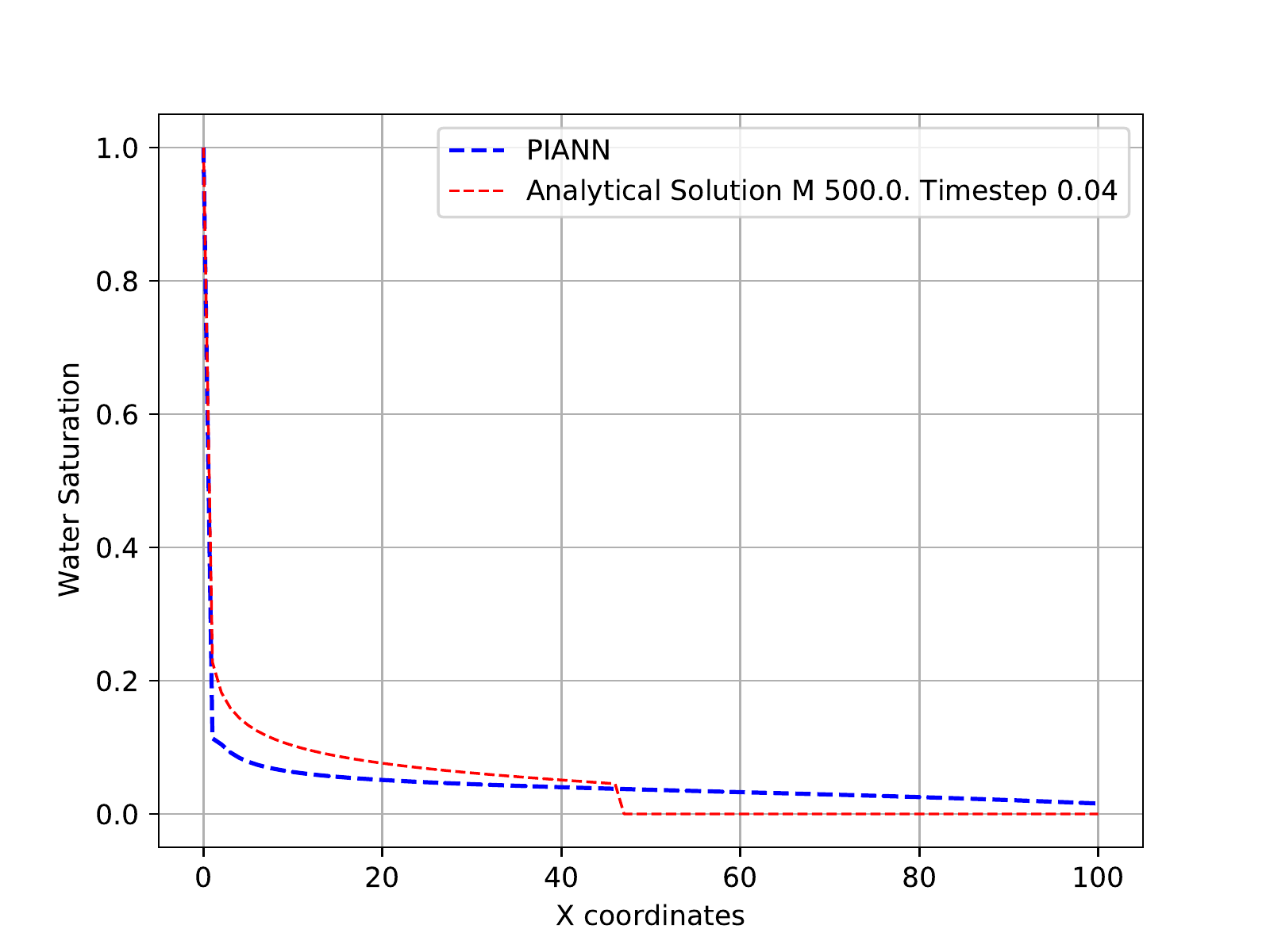}
     \includegraphics[width=0.28\linewidth]{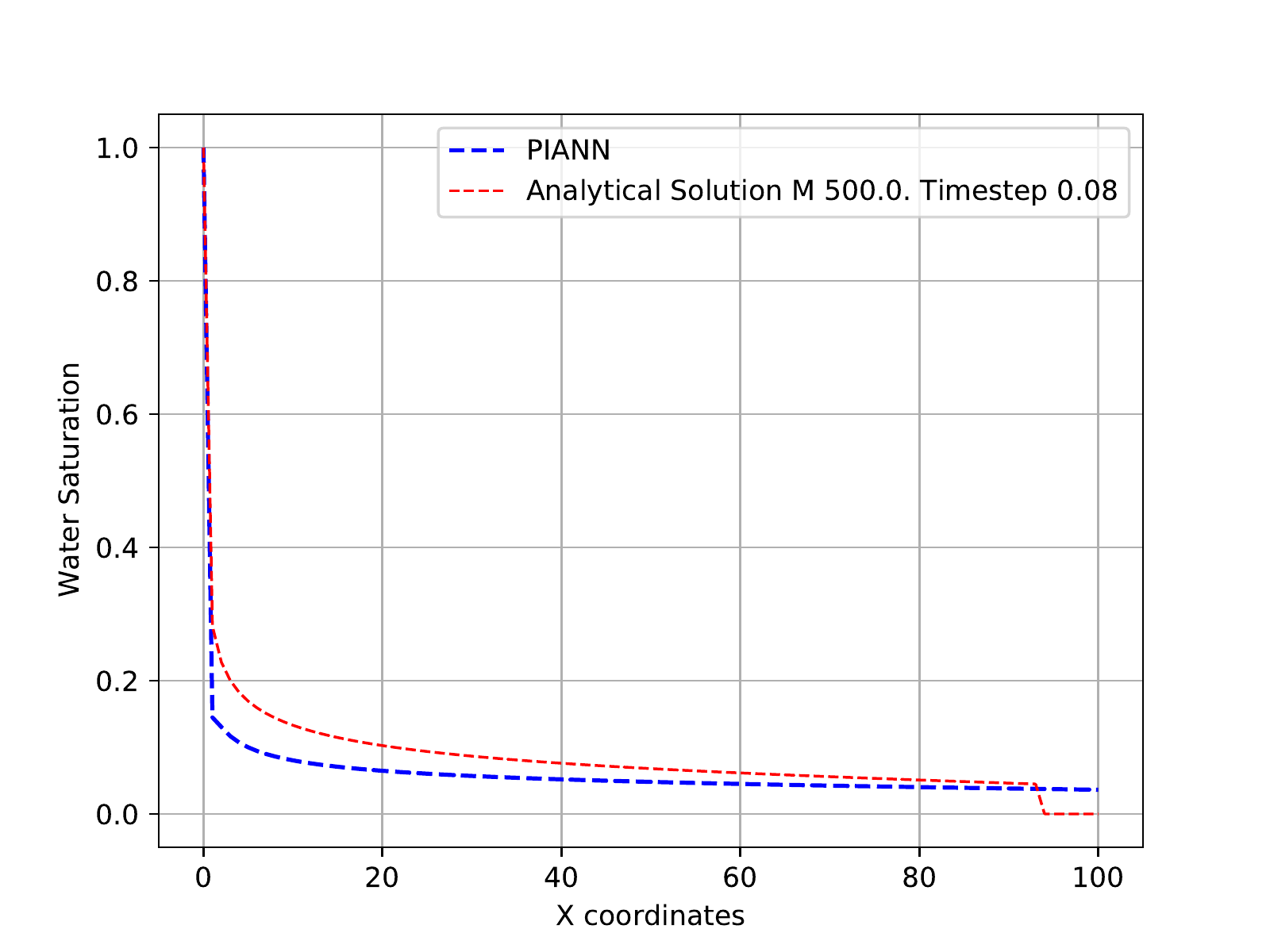}
     \includegraphics[width=0.28\linewidth]{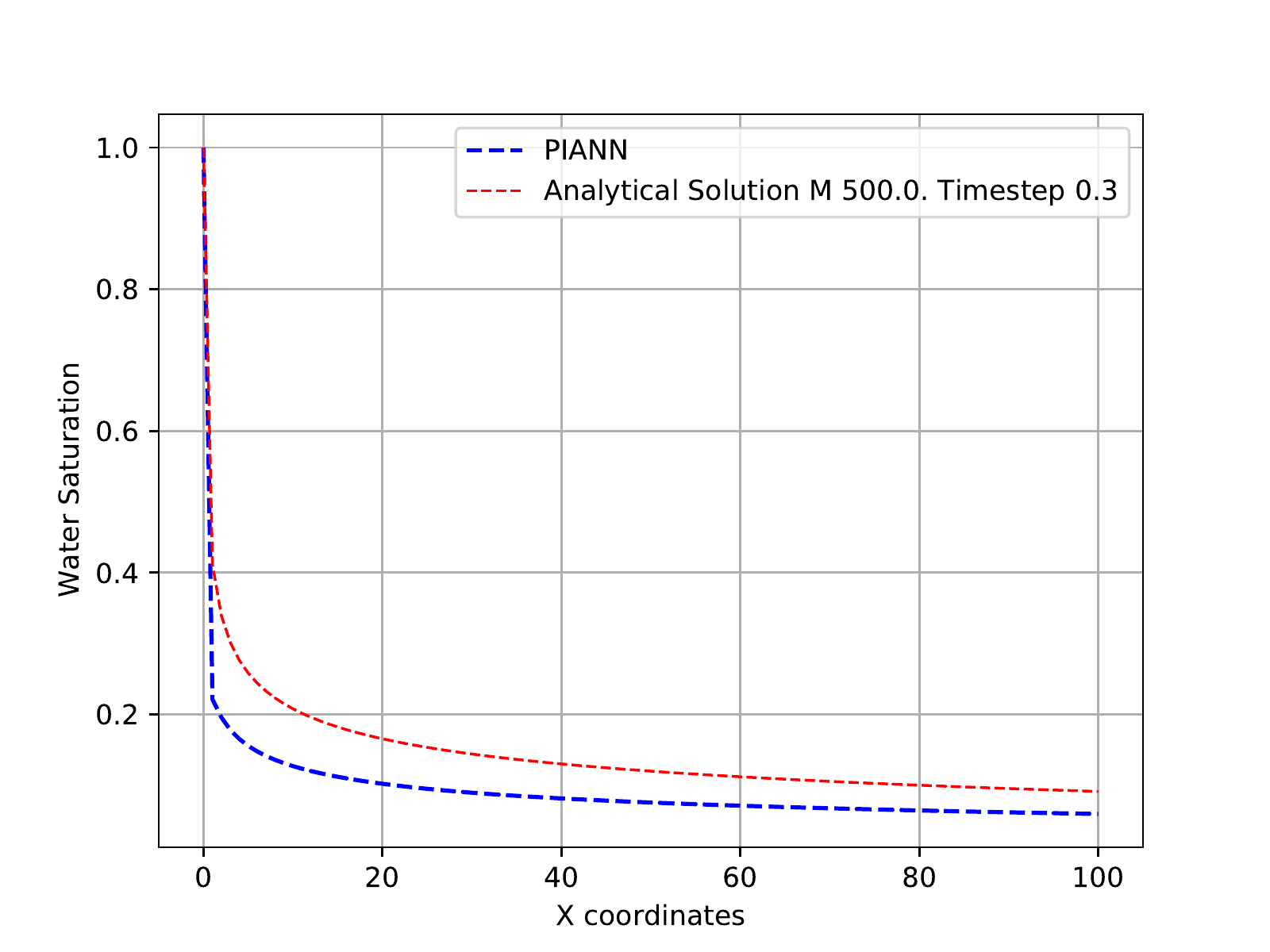}
     
     \caption{Top and bottom rows correspond to $ M = 140$ and $ M = 250 $ and $M=500$  comparison of the predicted by the neural network and the exact solutions of the PDE, respectively. The columns from left to right, correspond to different time steps $t = 0.04$, $t = 0.08$ and $t=0.30$}
     \label{fig:extrapolation}
     
 \end{figure*}

We also test the PIANN to extrapolate solutions out of the range of the training set: $M=140$, $M=250$ and $M=500$. Figure \ref{fig:extrapolation} shows a degradation of the results when $M$ is far from the original training set. We observe that our method predicts the behavior of the shock for $M=140$. However, the shock is totally missed for $M=500$ and as such, retraining the model is recommended with higher values of $M$. It is important to note that the results show that our method is stable and converges for the different cases. 

We test how the neural residual error progresses based on different $\Delta t$ and $\Delta x$ resolutions. Results are shown in table \ref{tab:resolutions}, and demonstrate that our PIANN obtains smaller residuals when the resolution of the training set increases. However, we observe that changes in the residual are not highly significant. This is an advantage with respect to traditional numerical methods such as CFD, where smaller values of $\Delta t$ are necessary to capture the shock and guarantee convergence and stability. 

Finally, we have compared the results with central and upwind finite difference schemes for the term of the vector of fluxes. The first-order upwind difference introduces a dissipation error when applied to the residual of the Buckley-Leverett equation, which is equivalent to regularizing the problem via artificial diffusion. Figure \ref{fig:residuals_comparation} shows that both approaches present similar results respect to the analytical solution. The fact that both central and upwind differences yield similar predictions is important, because it suggests that the proposed PIANN approach does not rely on artificial dissipation for shock capturing.

\section{Discussion}\label{sec:conclusions}



In this work, we have introduced a new method to solve hyperbolic PDEs. We propose a new perspective by focusing on network architectures rather than on residual regularization. We call our new architecture a physics informed attention neural network (PIANN).

PIANN's novel architecture is based on two assumptions. First, correlations between values of the solution at all the spatial locations must be exploited, and second, the architecture has to be flexible enough to identify the shock and capture different behaviors of the solution at different regions of the domain. We have proposed an encoder-decoder GRU-based network to use the most relevant information of the fully encoded information, combined with the use of an attention mechanism. The attention mechanism is responsible for identifying the shock location and adapting the behavior of the PIANN model.

Unlike previous methods in the literature, the loss function of PIANNs is based solely on the residuals of the PDE, and the initial and boundary conditions are introduced in the architecture. These are stronger constraints than the ones enforced by previous methods, since we do not allow room for learning error on the initial or boundary conditions. As a result, PIANN's training aims only at minimizing the residual of the PDE; no hyperparameters are needed to control the effect of initial and boundary conditions on the solution.

We have applied the proposed methodology to the non-concave flux Buckley-Leverett problem, which has hitherto been an open problem for PINNs. The experimental results support the validity of the proposed methodology and conclude that: i) during training, the residuals of the equation decrease quickly to values smaller than $10^{-4}$, which means that our methodology is indeed solving the differential equation, ii) the attention mechanism automatically detects shock waves of the solution and allows the PIANN to fit to the different behaviors of the analytical solution, and iii) the PIANN is able to interpolate solutions for values of the mobility ratio $M$ inside the range of training set, as well as to extrapolate when the value of $M$ is outside the range. However, we observe that if $M$ is too far away from range of the training set, the quality of the solution decreases. In that case, a retraining of the network is recommended. iv) We observe that the residuals decrease when the resolution of the training set increases. However, the change in the residuals is not highly significant. This is advantageous with respect to traditional numerical methods where small values of $\Delta t$ are needed to capture the shock and guarantee convergence and stability.

In conclusion, the proposed methodology is not confined by the current limitations of deep learning for solving hyperbolic PDEs with shock waves, and opens the door to applying these techniques to real-world problems, such as challenging reservoir simulations or carbon sequestration. It is plausible that this method could be applied to model many processes in other domains which are described by non-linear PDEs with shock waves.



\begin{table}
\centering
\caption{Residual calculation for different resolution of $\Delta t$ and $\Delta x$ for $M=4.5$}\label{tab:resolutions}
\begin{tabular}{lrrr}
Resolution & Residual Error  \\
\midrule
$\Delta x=1*10^{-2}$ $\Delta t=1*10^{-2}$ & $1*10^{-4}$  \\
$\Delta x=5*10^{-3}$ $\Delta t=5*10^{-3}$  & $9*10^{-5}$   \\
$\Delta x=1*10^{-3}$ $\Delta t=1*10^{-3}$  & $8.7*10^{-5}$ \\
\bottomrule
\end{tabular}
\end{table}

\bibliographystyle{unsrt}






\end{document}